\documentclass[letterpaper, 10 pt, journal, twoside]{ieeetran}

\IEEEoverridecommandlockouts

\usepackage{amsmath}
\usepackage{amssymb}

\DeclareMathOperator*{\argmin}{arg\,min}

\usepackage{algorithm}
\usepackage{algorithmic}

\usepackage{graphicx}
\usepackage{subfig}
\usepackage{array,booktabs,arydshln}

\usepackage{xcolor}
\usepackage{cite}

\definecolor{green}{RGB}{3,112,15}

\newcommand{\eg}{\textit{e.g.,}}

\title{Learning Inverse Kinodynamics for \\Accurate High-Speed Off-Road Navigation on Unstructured Terrain
}

\author{Xuesu Xiao$^{1}$, Joydeep Biswas$^{1}$,  and Peter Stone$^{1, 2}$
\thanks{Manuscript received: February, 24, 2021; Revised April, 25, 2021;
Accepted June, 5, 2021.}
\thanks{This paper was recommended for publication by Editor Dan Popa upon
evaluation of the Associate Editor and Reviewers' comments. 
}
\thanks{$^{1}$Xuesu Xiao, Joydeep Biswas, and Peter Stone are with Department of Computer Science, The University of Texas at Austin, Austin, TX 78712. {\tt\small \{xiao, joydeepb, pstone\}@cs.utexas.edu}}
\thanks{$^{2}$Peter Stone is with Sony AI.}%
\thanks{Digital Object Identifier (DOI): see top of this page.}
}

\begin{document}
\maketitle

\begin{abstract}
This paper presents a learning-based approach to consider the effect of unobservable world states in kinodynamic motion planning in order to enable accurate high-speed off-road navigation on unstructured terrain. 
Existing kinodynamic motion planners either operate in structured and homogeneous environments and thus do not need to explicitly account for terrain-vehicle interaction, or assume a set of discrete terrain classes.   
However, when operating on unstructured terrain, especially at high speeds, even small variations in the environment will be magnified and cause inaccurate plan execution.  
In this paper, to capture the complex kinodynamic model and mathematically unknown world state, we learn a kinodynamic planner in a data-driven manner with onboard inertial observations. 
Our approach is tested on a physical robot in different indoor and outdoor environments, enables fast and accurate off-road navigation, and outperforms environment-independent alternatives, demonstrating 52.4\% to 86.9\% improvement in terms of plan execution success rate while traveling at high speeds. 

\end{abstract}

\section{Introduction}
\label{sec::intro}
\IEEEPARstart{C}{urrent}
mobile robot navigation methods can navigate a robot from one point to another safely and reliably in structured and homogeneous environments~\cite{quinlan1993elastic, fox1997dynamic}, such as indoor hallways or outdoor paved surfaces. These consistent environments allow the robots to use simple kinodynamic motion planners independent of the environment, thanks to the limited environment disturbances and stochasticity. 

Dating back at least to DARPA's Grand Challenge~\cite{seetharaman2006unmanned} and LAGR (Learning Applied to Ground Vehicles)~\cite{jackel2006darpa} program, researchers have also looked into applying autonomous navigation in unstructured outdoor environments. Challenges arise from multiple fronts in those natural spaces, but most off-road navigation work focused on perception, e.g., detecting natural obstacles~\cite{jackel2006darpa}, classifying underlying terrain types~\cite{bai2019three, shi2020laplacian, mei2019comparative}, or building semantic maps~\cite{maturana2018real, wolf2020advanced, sharma2019semantic}. For motion control, most off-road robots simply travel at low speeds to minimize uncertainty and to maximize safety~\cite{rabiee2019friction, tian2014control, rogers2012aiding, seyr2006proprioceptive}. 

While recent advances in deep learning provide roboticists with a different avenue to investigate those perception problems in off-road navigation, researchers also started to combine perception, planning, and motion control using end-to-end learning in unstructured environments~\cite{muller2006off, giusti2015machine}. These systems do not require a heavily-engineered navigation pipeline, and can react to natural environments in a data-driven approach. Although these methods can enable successful navigation, they are data-intensive and and generally do not lead to better navigation than their classical counterparts. 

\begin{figure}
  \centering
  \includegraphics[width=1\columnwidth]{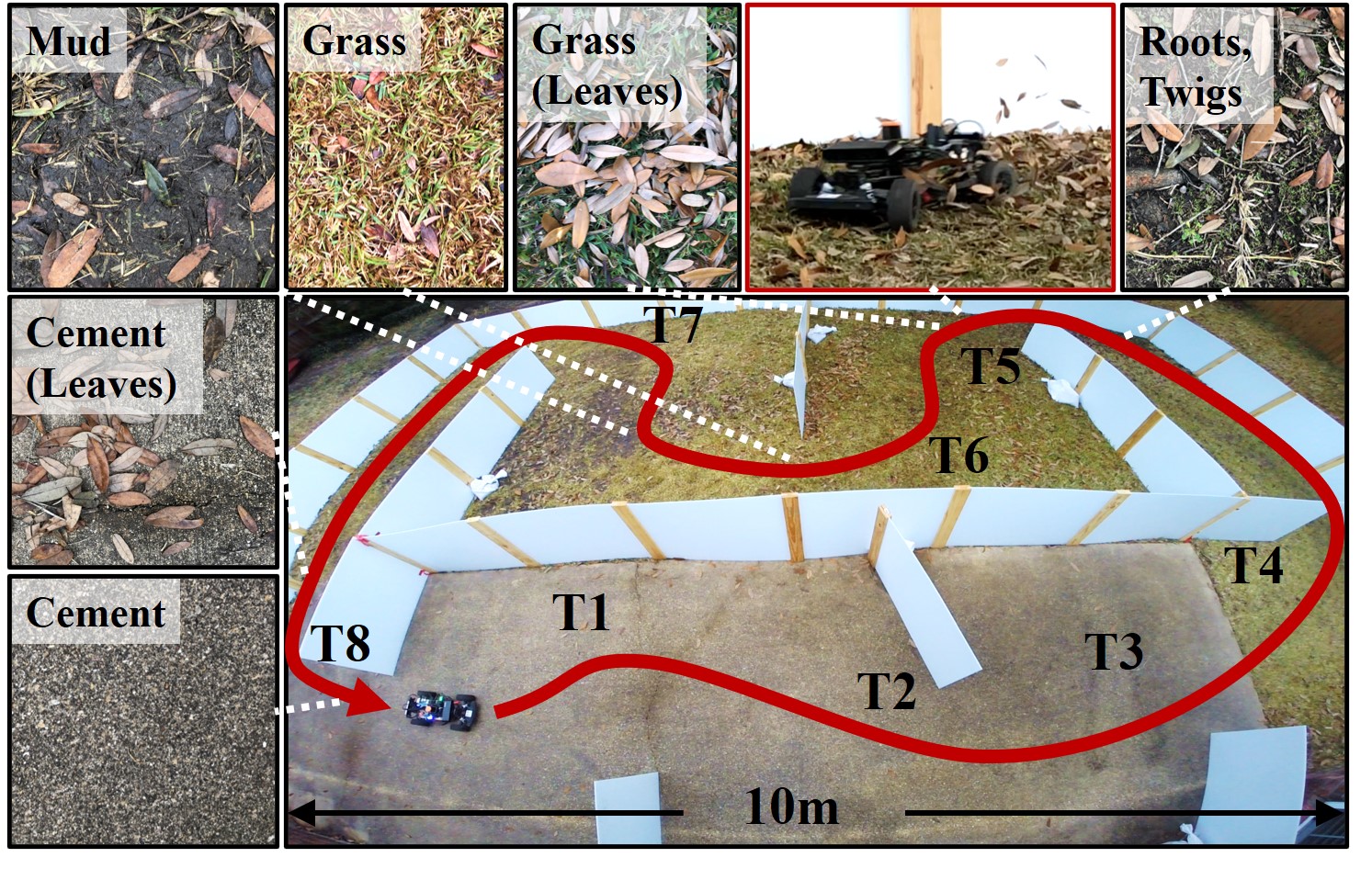}
  \caption{The UT Automata scale 1/10th autonomous vehicle drives an eight-turn (T1--T8 on the red path) outdoor race track with unstructured terrain. Close-ups of some terrain (black inserts) are shown, including cement, mud, grass, which are covered by leaves, stalks, and/or twigs at different densities. For high-speed terrain-aware navigation, the robot has to interact with these unstructured terrain (red insert). }
  \label{fig::backyard}
\end{figure}

Focusing on the motion control side of off-road navigation on unstructured terrain, the contribution of this paper is to use learning to capture the effect of complex and unknown environmental factors on the robot's low-level kinodynamic model. 
We use onboard inertial observation to encode environmental factors and learn an inverse kinodynamic model to produce fast and accurate low-level control inputs. Using our method, even with extensive disturbances caused by high-speed terrain-vehicle interaction, the robot is still able to perform fast and accurate off-road navigation (Fig. \ref{fig::backyard}), compared to a kinodynamic model based on ideal assumptions and a learned model that does not consider environmental factors. 

\section{Related Work}
\label{sec::related}

In this section, we review related off-road navigation work in terms of perception and motion control. 

\subsection{Off-road Perception}
The first challenge arises from off-road navigation is perception. In unstructured off-road environments, perception is no longer simply in the geometric sense (e.g., free vs. obstacle), but also requires semantic information (e.g., grass vs. mud). 
A plethora of research in terrain classification has leveraged vibration-based signals to classify terrain types~\cite{bai2019three, shi2020laplacian, mei2019comparative}. Vision-based sensors, e.g., LiDAR and camera, combined with current deep learning methods, have also been used to build semantic maps~\cite{maturana2018real, wolf2020advanced, sharma2019semantic}. 
These perception methods assign costs to discrete terrain classes for planning, but do not consider robot's kinodynamic model when moving on these terrain. Our work does not distinguish among discrete terrain classes and uses observations collected during interactions with different terrain to enable fast and accurate kinodynamic planning. 

\subsection{Off-road Motion Control}
Although research thrust for off-road navigation has been primarily focused on perception, roboticists have also investigated off-road navigation from the motion control side. Many wheel slip models~\cite{rabiee2019friction, tian2014control, rogers2012aiding, seyr2006proprioceptive} have been developed and used to design controllers to minimize slip. Most of these models treat slip only as a function of the vehicle kinematics. But to achieve high-speed off-road navigation, slip is inevitable and also highly dependent on the underlying terrain. 

Researchers have also used machine learning for motion control in off-road navigation. A recent survey~\cite{xiao2020motion} pointed out that learning is most efficient when targeting at navigation components, e.g., learning local planners~\cite{wangagile, xiao2021agile, xiao2020toward, liu2020lifelong, chiang2019rl, kontoudis2019kinodynamic}, costmaps~\cite{wigness2018robot, richter2017safe}, or planner parameters~\cite{xiao2021appl, xu2020applr, wangapple, wang2020appli, xiao2020appld}. 
Research on using learning for motion control in off-road scenarios is scarce. Pan, et al.~\cite{pan2020imitation} enabled high-speed navigation with end-to-end imitation learning from RGB input in a closed circular dirt track. The expert demonstrator is a model predictive controller with access to high-precision sensors including GPS-RTK. The end-to-end learning approach most likely does not generalize well to other terrain and tracks. Aiming at a variety of terrain, Siva, et al.~\cite{siva2019robot} used imitation learning from human demonstration to navigate five discrete terrain types (concrete, grass, mud, pebble, rock). 
In contrast, our method only targets at learning a kinodynamic model and can navigate any global path. We also do not intentionally separate terrain into discrete types, and treat different terrain characteristics in a continuous manner. Reinforcement learning has also shown potential in learning motion control for navigation, but at the cost of extensive training overhead: Brunnbauer, et al. \cite{brunnbauer2021model} took 8 million time steps (37 hours of interactions) to learn a simple \emph{kinematic} model. In comparison, our approach only requires 30 minutes to learn a \emph{kinodynamic} model, which would require far more training by such reinforcement learning approaches.

\section{Learning Inverse Kinodynamic Models}
\label{sec::approach}
Most navigation systems either assume kindodynamic models to be independent of the environment, or that there exists a discrete set of environment classes~\cite{siva2019robot}, e.g., one model for paved terrain and another for grass. In this work, we relax these assumptions by learning a single continuous inverse kinodynamic model that accounts for environmental factors across different terrain types without having to perform discrete terrain classification, or analytic modelling. The learned model takes as input inertial observations that make the impact of environmental factor on kinodynamic motion observable (e.g., how bumpiness from gravel will result in understeer at high speeds). This inverse kinodynamic model is learned in a data-driven manner.

\subsection{Problem Formulation}
Given vehicle state $x$,
control input $u$, and world state $w$, the state dynamics and observation $y$ are given by 
\begin{equation} \label{eqn::dynamics}
\dot{x}  = f(x, u, w), 
\qquad y  = g(x, w),
\end{equation}
where $f(\cdot, \cdot, \cdot)$ is the system's forward kinodynamic function, while $g(\cdot, \cdot)$ is the observation function. Note that in most cases, $w$ is not directly observable and cannot be easily modeled. 
A navigation planner generates a global plan $\Pi: [0,1] \rightarrow X$ mapping from a unitless progress variable $s\in [0,1]$ to planned vehicle state $x \in X$, incorporating both global (\eg{} traversable map) and local (\eg{} sensed obstacles) information to take the robot from the start state $\Pi(0)$ to the goal state $\Pi(1)$. A projection operator $\rho: X\rightarrow [0,1]$ maps the robot state $x$ (e.g., from localization) to infer the progress variable $s$ (i.e., the robot’s progress along the global plan so far), such that the closest state in the plan to a robot state $x$ is $\Pi(\rho(x))$. For simplicity of notation, we represent the projected state at any time as $x_\Pi=\Pi(\rho(x))$. We also omit the explicit time-dependence of variables $x(t)$, $u(t)$, and $y(t)$, denoting them simply as $x$, $u$, and $y$. The objective of our controller $u$ is thus to minimize the total navigation time $T$ while following the plan precisely, as represented by the joint cost function
\begin{align}
    J = T + \gamma \int_0^T ||x(t) - x_\Pi(t)||^2 dt.
    \label{eqn::objective}
\end{align}
Here, $\gamma$ is a hyperparameter that trades total navigation  time for execution accuracy.

We formulate the solution to this optimal control problem as a receding-horizon controller $u^*$ over a unitless progress horizon (along the global plan) $\Delta$ and corresponding time-horizon $\Delta t$ such that the control input drives the robot state from $x$ to the receding horizon plan state $\Pi(\rho(x)+\Delta)$ over time-period $\Delta t$:
\begin{align}
    &u^* = \argmin_u \left( \Delta t + \gamma \left\lVert \Delta x_\Pi - \int_0^{\Delta t} f(x, u, w)dt \right\rVert^2 \right),  \nonumber \\ 
    &\Delta x_\Pi = \Pi(\rho(x)+\Delta) - x, \label{eqn::controller}
\end{align}
where $\Delta x_\Pi$ is the state change between the receding horizon plan and current vehicle state.\footnote{In general, the minus signs in Eqn. \ref{eqn::controller} is the generalized difference operator $\ominus$ over Special Euclidean Group $\mathrm{SE}(n)$ and the corresponding Lie Algebra.} The optimal control $u^*$ can be solved using the receding horizon inverse kinodynamic model $f^{-1}$ as
\begin{align}
    u^* = f^{-1}(\Delta x, x, w), \label{eqn::ikd}
\end{align}
that takes as input the desired relative state change $\Delta x$, the current robot state $x$, and world state $w$. Unfortunately, it is hard to express $f^{-1}$ accurately via analytical models, and even if it could be expressed accurately,  computing $u^*$ is error-prone since the world state $w$ is not directly observable.

\subsection{Learning Inverse Kinodynamics}
In this work, in order to enable fast and accurate navigation under the influence of different terrain interactions from the world state $w$, we adopt a data-driven approach to capture the effect of $w$. Specifically, we introduce the function 
\begin{equation}
    f^{+}_\theta(\Delta x, x, y) \approx f^{-1}(\Delta x, x, w), 
\end{equation}
parameterized by $\theta$, as an approximation for the original receding horizon inverse kinodynamic function (superscript $^+$ denotes pseudo inverse). 
The key insight in this approximation is that the impact of $w$ on $u^*$ becomes predictable given observations $y$ related to high-speed terrain-aware navigation---in our case we use onboard inertial sensing to capture speed-dependent terrain interaction. 
To learn $f^{+}_\theta$, a training dataset $\mathcal{T}$ with $N$ samples
\[
\mathcal{T} = \{\langle \Delta x^i, x^i, u^i, y^i\rangle_{i=1}^{N}\}
\]
 is desired, using the optimal but unknown receding horizon inverse kinodynamic function $u^i=f^{-1}(\Delta x^i, x^i, w^i)$ and observation function $y^i = g(x^i, w^i)$ from Eqns.~\ref{eqn::dynamics} and \ref{eqn::ikd}. Unfortunately, we neither know $f^{-1}$, nor do we know the world states $w^i$. However, we do have access to $f$ as a black-box function via real-world execution: we can simply pick arbitrary sample controls $u^i$ at corresponding starting states $x^i$, and observe the resulting state change $\Delta x^i$ after the chosen receding horizon $\Delta t$, including the impact of the unknown  $w^i$:
 \[
 \Delta x^i = \int_0^{\Delta t} f(x^i,u^i,w^i)dt.
 \]
Thus, the original chosen control $u^i$ is 
the control\footnote{Here, we assume the optimal control that follows the global path (second term in Eqn. \ref{eqn::objective}) also implicitly minimizes navigation time (first term).}
for the resulting state change $\Delta x^i$ from the original state $x^i$, for the corresponding but unknown (and hence unrecorded) world state $w^i$. Along with the corresponding sensor observation $y^i$, we generate each sample $i$ for dataset $\mathcal{T}$.
 To ensure that the learned parameters $\theta$ of $f^{+}_\theta$ approximate $f^{-1}$ accurately at states that the robot will encounter during execution, $\mathcal{T}$ must include representative samples for $x$, $u$, and $y$.
 With the collected dataset $\mathcal{T}$, we formulate deriving $f^{+}_\theta(\cdot, \cdot, \cdot)$ as a learning problem by minimizing a supervised loss: 
\begin{equation}
    \begin{split}
    \theta^* &= \argmin_{\theta} \sum_{(\Delta x^i, x^i, y^i) \in \mathcal{T}} \lVert f^{-1}(\cdot, \cdot, \cdot) - f^{+}_\theta(\Delta x^i, x^i, y^i)\rVert_H \\
    &= \argmin_{\theta} \sum_{(u^i, \Delta x^i, x^i, y^i) \in \mathcal{T}} \lVert u^i - f^{+}_\theta(\Delta x^i, x^i, y^i)\rVert_H, 
    \end{split}
    \label{eqn:bc}
\end{equation}
where $||v||_H = v^THv$ is the norm induced by positive definite matrix $H$, used to weigh the learning loss between the different dimensions of the control input $u^i$. We represent $f^{+}_\theta(\cdot, \cdot, \cdot)$ as a neural network and can therefore use gradient decent to find an approximately optimal $\theta^*$. In this work, we collect raw 6-DoF readings from an onboard IMU, and construct $y$ by feeding inertial data through an autoencoder to encode relevant terrain-vehicle interaction at different driving speeds. More details are provided in Sec. \ref{sec::experiments}. 

\subsection{Online Execution}
The learned inverse kinodynamic model $f^{+}_{\theta^*}(\cdot, \cdot, \cdot)$ provides a means to approximately account for $w$ using the onboard observation $y$ for fast, terrain-aware, and precise navigation. At each time step $t$ during online execution,
we compute the desired change of state $\Delta x_\Pi = \Pi(\rho(x)+\Delta) - x$ with $x$ from localization, the projection operator $\rho(\cdot)$, projection horizon $\Delta$, and global plan $\Pi(\cdot)$.
Along with onboard observation $y$ and current vehicle state $x$, we use the learned inverse kinodynamic model $f^{+}_{\theta^*}$ to produce system control input: 
\begin{equation}
    u(t) = f^{+}_{\theta^*}(\Delta x_\Pi, x, y), 
\end{equation}
and repeat this process for every time step.

\section{Experiments}
\label{sec::experiments}
In this section, we present experimental results using a learned inverse kinodynamic model $f^{+}_{\theta^*}(\Delta x, x, y)$, which considers unobservable world state $w$ by taking $y$ as input, and can precisely track different global plans by different $\Pi$. 

We denote the baseline forward and inverse kinodynamic functions, which do not consider world state $w$, as $f_B(x, u)$ and $f^+_B(\Delta x, x)$, respectively. As an ablation study to test the effectiveness of capturing the world state $w$ with $y$, we also learn an ablated inverse kinodynamic model $f^+_{A\phi^*}(\Delta x, x)$, which is parameterized by $\phi^*$ and does not take observation $y$ as input to represent $w$. 
Note that the baseline represents classical model-based local planners \cite{fox1997dynamic}, while the ablation is equivalent to existing learning-based local planners, e.g., Imitation Learning \cite{pfeiffer2017perception}. 
We show that the learned $f^+_{A\phi^*}$ can outperform the baseline $f^+_B$. Adding the learned observation $y$ as another input, $f^{+}_{\theta^*}(\Delta x, x, y)$ can outperform both  $f^+_{A\phi^*}(\Delta x, x)$ and $f^+_B(\Delta x, x)$. 

The learned inverse kinodynamic model for online execution is also agnostic to different online plans and unseen terrain. We show that the model learned through a global planner $\Pi_1$ (a randomly exploring policy driven by a human operator) can also generalize well to a complicated global planner $\Pi_2$ on an outdoor race track, and a simple global planner $\Pi_3$ on an indoor track. 
We also test the generalizability with respect to unseen world states through the encoded $y$ and $f^{+}_{\theta^*}(\Delta x, x, y)$ on unseen terrain. 

\subsection{Implementation}
\subsubsection{Robot Platform} 
\begin{figure}
\centering
\subfloat[]{\includegraphics[width=0.5\columnwidth]{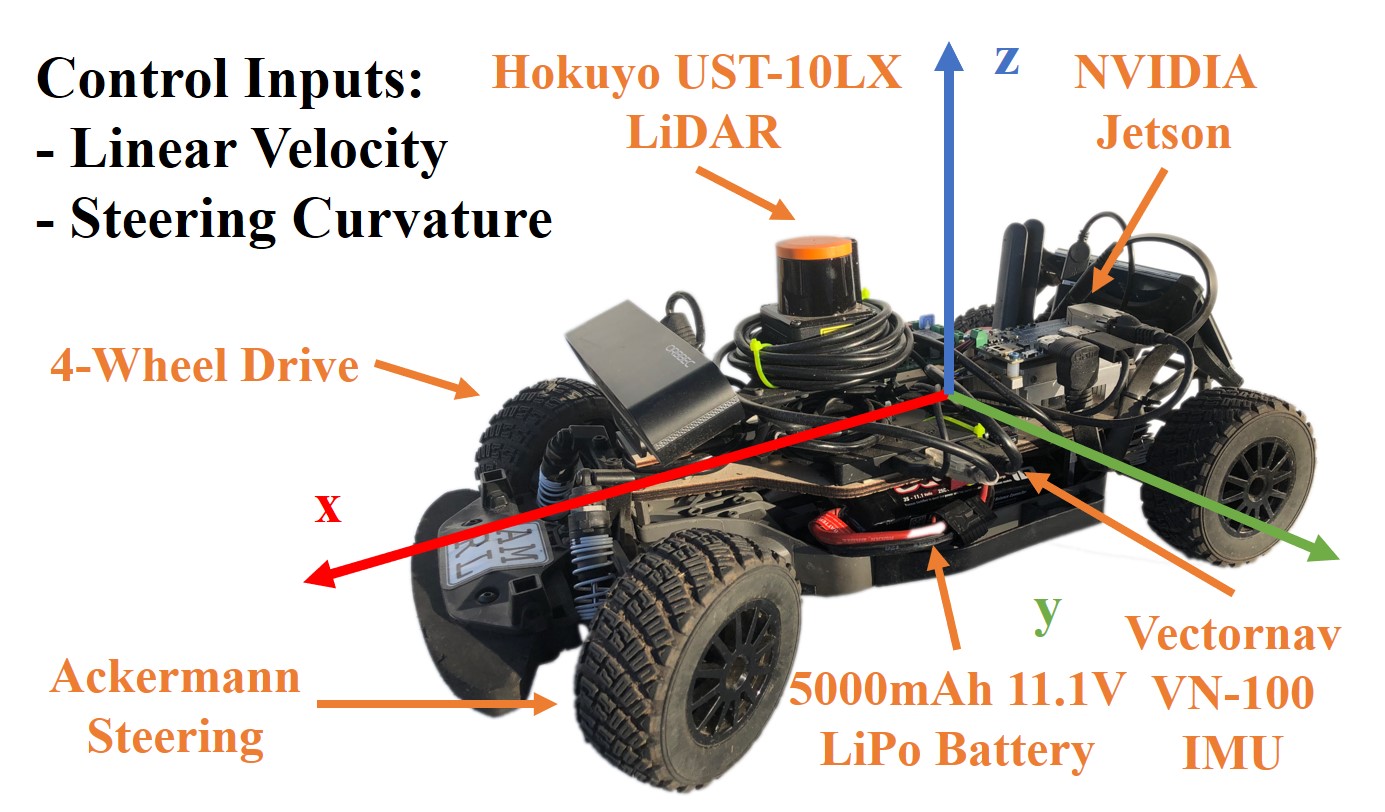}
\label{fig::ut_automata}}
\subfloat[]{\includegraphics[width=0.5\columnwidth]{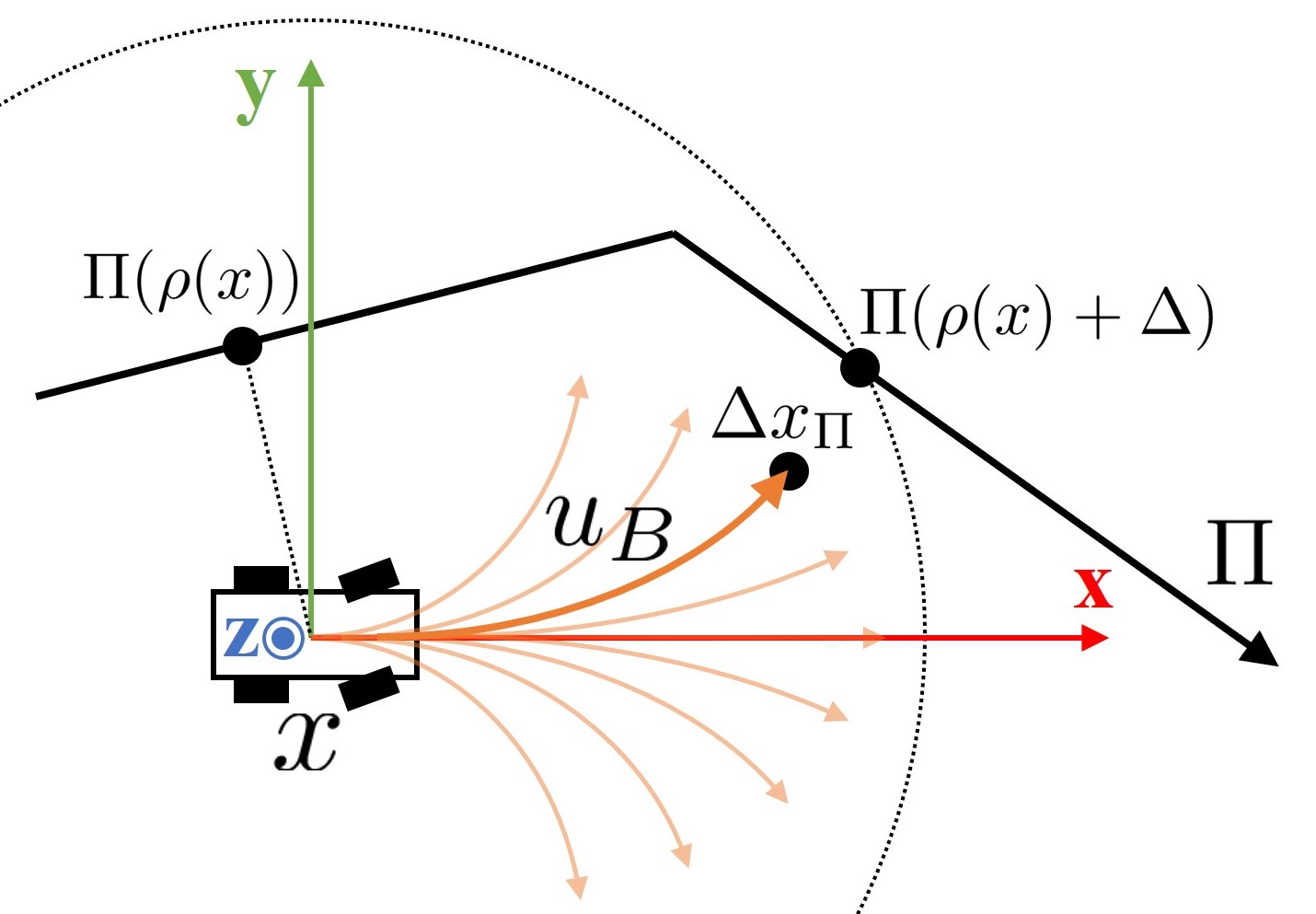}%
\label{fig::graph_nav_cut}}
\caption{(a): A UT Automata robot, scale 1/10th autonomous vehicle platform used in the experiments. (b): The sampling-based baseline approximates $\Delta x_\Pi$ by rolling out the optimal $u_B$ among 100 samples with $f_B$.}
\label{fig::ut_automata_and_graph_nav}
\vspace{-10pt}
\end{figure}

Our learned inverse kinodynamic model is implemented on a UT Automata robot, a scale 1/10th autonomous vehicle platform (Fig. \ref{fig::ut_automata}). The Ackermann-steering four-wheel drive vehicle is equipped with a 2D Hokuyo UST-10LX LiDAR for localization, a Vectornav VN-100 IMU for inertial sensing (200Hz), a Flipsky VESC 4.12 50A motor controller, and a Traxxas Titan 550 Motor. Although the platform has individual suspensions, the relatively short-travel suspensions are not specifically designed for off-road navigation. We specifically pick this platform because its small size and weight and the lack of designed off-road capability can maximize the difference in accuracy between navigating with and without the learned terrain-aware inverse kinodynamic model. The robot has a NVIDIA Jetson onboard, but only the CPU is used during deployment. 

\subsubsection{Environment} The environment comprises of cement, grass, and mud; and some patches are covered by different artifacts, such as leaves, stalks, and/or twigs, with different densities (Fig. \ref{fig::backyard}). For a small vehicle like the UT Automata robot, these artifacts can cause significantly different world states (Fig. \ref{fig::backyard} red insert). Note that the terrain also changes due to environmental factors such as sunlight, wind, and moisture, and is also affected by the robot's wheel and chassis. To minimize the effect of artifacts being pushed off the course during extended experiments, we frequently shuffle and redistribute the artifacts. We do not specify discrete terrain classes and treat the terrain characteristics in a continuous manner. 

\subsubsection{Model Implementation} During autonomous navigation, the robot uses Episodic Non-Markov Localization (ENML)~\cite{biswas2017episodic} with a pre-built map of the environment to derive vehicle state $x$. A global planner $\Pi$ includes a pre-generated global path for the robot to follow and uses line-of-sight control (similar to~\cite{xiao2017uav, xiao2015locomotive}) to generate desired receding horizon plan state $\Pi(\rho(x)+\Delta)$ on the global path 1m away from the robot. 
In a model predictive control manner, the robot uses the baseline forward kinodynamic function $f_B$ and samples candidate velocity and curvature control inputs $u\in U$ evenly distributed within a physically-feasible window to jointly find the desired state change $\Delta x_\Pi$ and control input $u_B$ (shown in Fig. \ref{fig::graph_nav_cut}).
More specifically, to compute control input, the baseline inverse kinodynamic model $f^+_B$ produces the curvature input, which results in the desired $\Delta x_\Pi$ and drives the robot as close to $\Pi(\rho(x)+\Delta)$ as possible: 
\begin{equation}
\begin{split}
    u_B & = f^+_B(\cdot, \cdot)\\
    & = \argmin_{u}\lVert\Pi(\rho(x)+\Delta) - \int_{t=0}^{\Delta t} f_B(x, u)dt \rVert,
\end{split}
\end{equation}
for the second term in Eqn. \ref{eqn::objective}, and selects the fastest possible velocity for the first term $T$, considering the robot's acceleration limit and a safety distance to decelerate in case of obstacles.

For our learned ablated and final model, $f^+_{A\phi^*}$ and $f^{+}_{\theta^*}$, we utilize the $\Delta x_\Pi$ from the baseline kinodynamic model (corresponds to $u_B$), but instead of using the baseline's control input, we query our learned models to produce 
$u = f^+_{A\phi^*}(\Delta x_\Pi, x)$
or
$u = f^{+}_{\theta^*}(\Delta x_\Pi, x, y)$. In practice, we use the baseline control input $u_B = \{v, c\}$ (linear velocity and steering curvature) to represent the desired state change rate $\Delta x_\Pi$.

\subsubsection{Data Collection} To collect training data, the robot is teleoperated with a joystick in an open environment with linear velocity $v \in [0, 3]\mathrm{m/s}$ and steering curvature $c \in [-1.35, 1.35]\mathrm{m^{-1}}$ for 30 minutes (24418 data points). The teleoperator randomly varies both linear velocity and steering curvature ($\Pi_1$). 
In our specific implementation, the robot reasons in the robot frame and therefore the state $x^i$ in the training trajectory $\mathcal{T} = \{\langle \Delta x^i, x^i, u^i, y^i\rangle_{i=1}^{N}\}$ becomes the origin in the robot frame. 
The ground truth $\Delta x^i$ is represented as real $\{v^i_r, c^i_r\}$, where $v^i_r$ is from vehicle odometry and $c^i_r = \omega^i_r/v^i_r$ ($\omega^i_r$ is the sensed angular velocity around the vertical $z$ axis from the IMU). Currently, we take $v^i_r$ from wheel odometry only, which can be further improved by adding visual, point cloud, and/or inertial information in future work. The commanded control input $u^i=\{v^i_c, c^i_c\}$ is recorded from joystick input. For $y$, we collect the 6-DoF raw IMU signal, including 3-DoF accelerometer and 3-DoF gyroscope, as a sliding history window. 

\subsubsection{Network Architecture} 
\begin{figure}
  \centering
  \includegraphics[width=0.9\columnwidth]{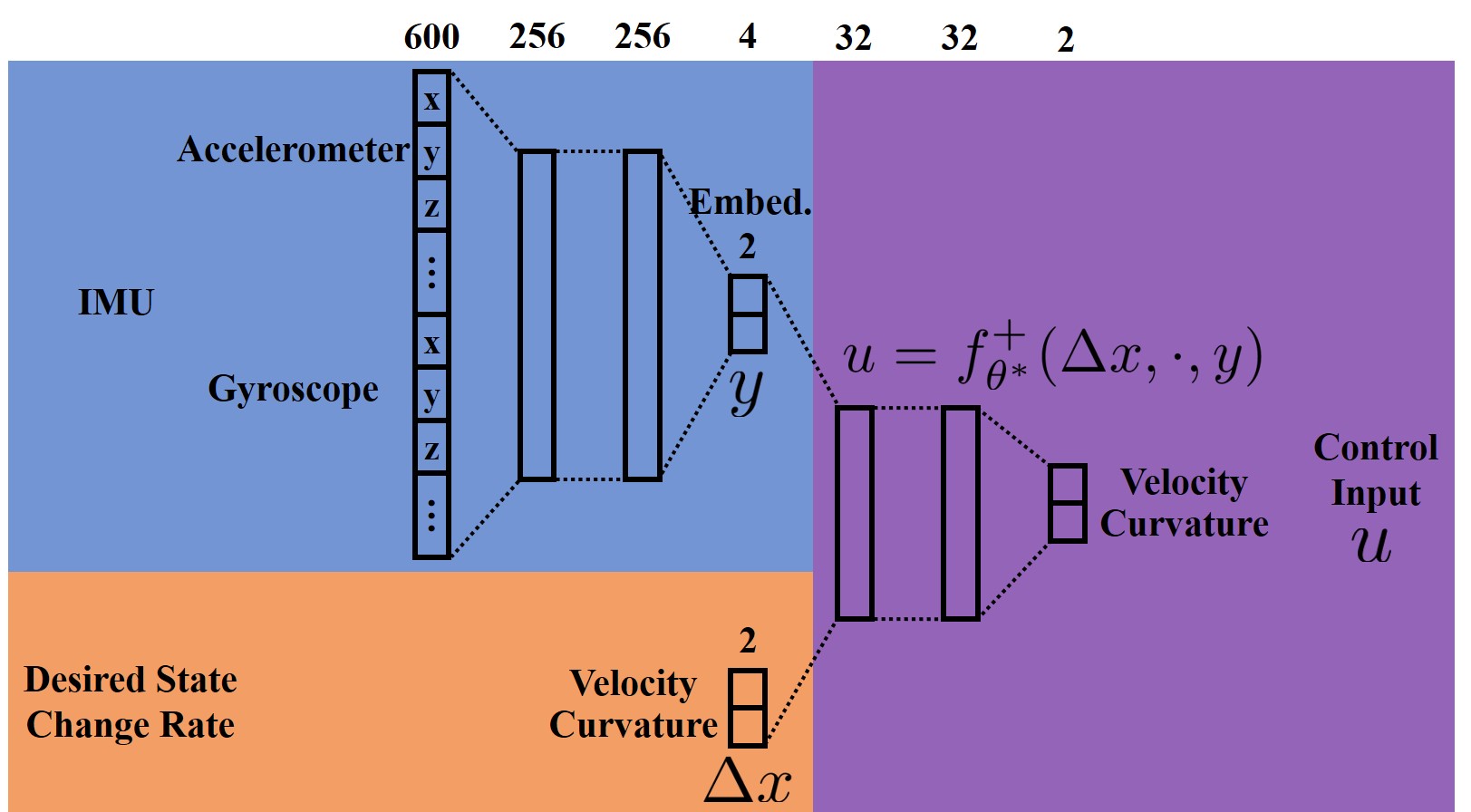}
  \caption{Neural Network Architecture. Input: blue IMU encoder and orange desired state change (desired velocity and curvature in practice); Output: purple learned function approximator $f^{+}_{\theta^*}$ as the inverse kinodynamic model. }
  \label{fig::nn}
  \vspace{-10pt}
\end{figure}

As a function approximator for $f^{+}_{\theta^*}$, we use a two-layer neural network with 32 neurons each layer (shown in purple in Fig. \ref{fig::nn}) and we empirically show that such a small network is very efficient to train and suffices to improve off-road navigation. 
The neural network takes realistic/desired $\{v^i_r, c^i_r\}$ (as a proxy for $\Delta x^i$, Fig. \ref{fig::nn}, orange) and observation $y$ as input, and outputs to-be-commanded control input $u^i=\{v^i_c, c^i_c\}$. 
For observation $y$, we concatenate the last 100 IMU readings (0.5s) into a 600-dimensional vector, and feed it into two 256-neuron layers as an autoencoder (Fig. \ref{fig::nn}, blue). 
The final embedding for $y$ is a two dimensional vector, then concatenated with $\{v^i_r, c^i_r\}$, and finally trained in an end-to-end fashion. 
The entire network architecture is shown in Fig. \ref{fig::nn}. For the ablated model $f^+_{A\phi^*}$, the two-dimensional $y$ embedding is removed (only the orange and purple components remain). Training both models takes less than five minutes on a NVIDIA GeForce GTX 1650 laptop GPU. During runtime, the trained model is used onboard the robot's Jetson CPU with \texttt{libtorch}. 

\begin{figure*}
\centering
\subfloat[1.6m/s]{\includegraphics[width=0.4\columnwidth]{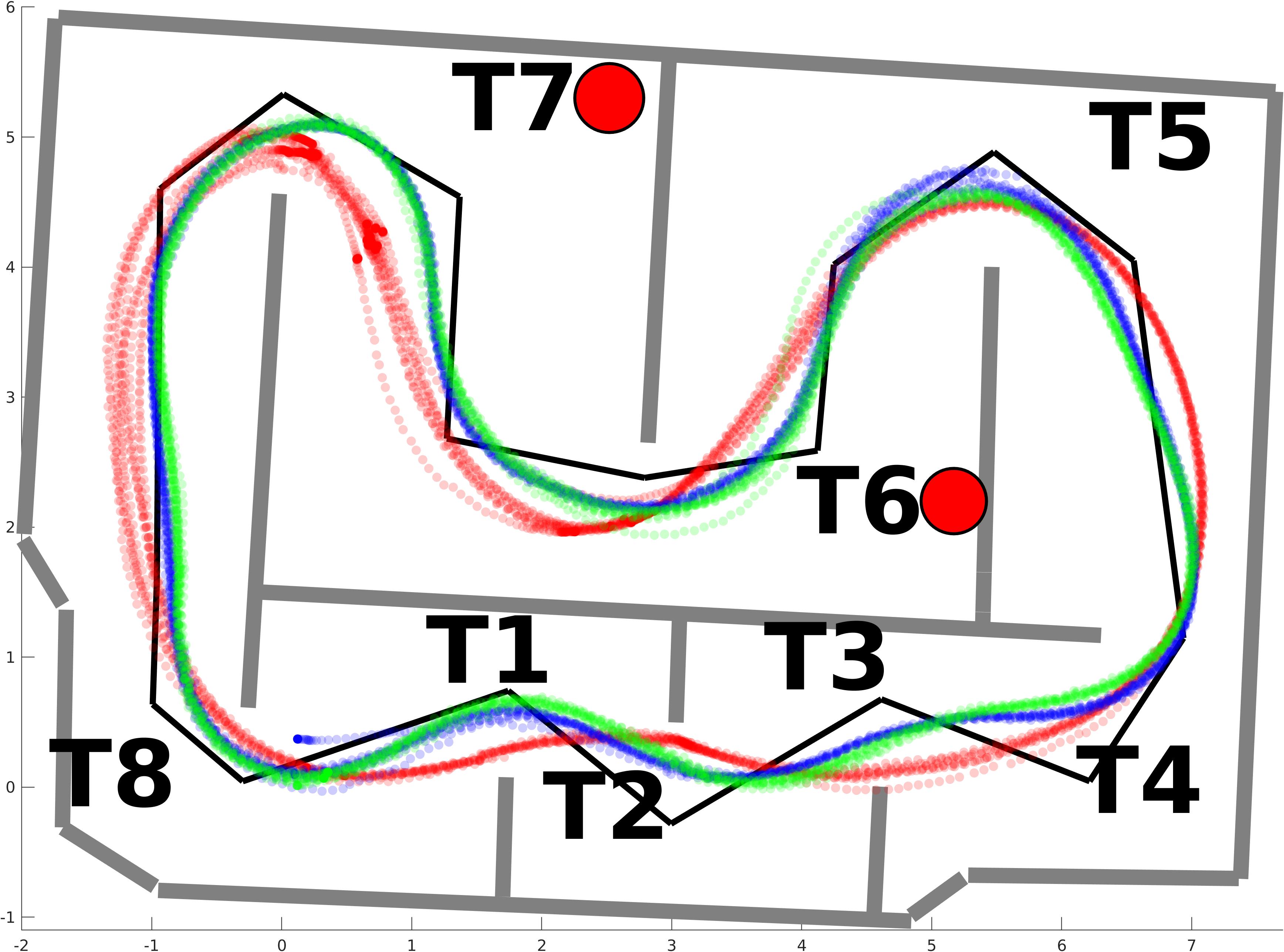}%
\label{fig::1.6}}
\subfloat[1.7m/s]{\includegraphics[width=0.4\columnwidth]{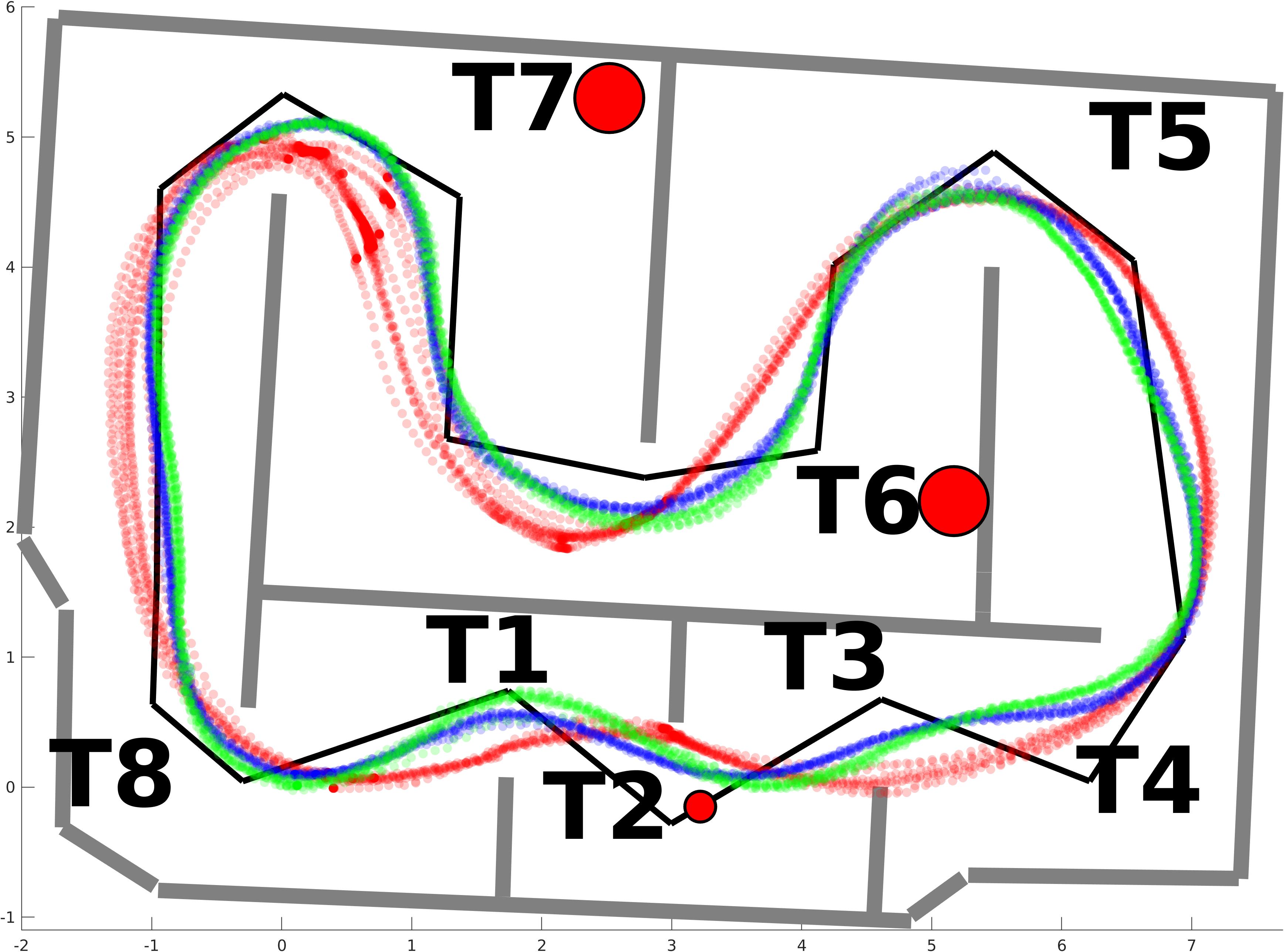}%
\label{fig::1.7}}
\subfloat[1.8m/s]{\includegraphics[width=0.4\columnwidth]{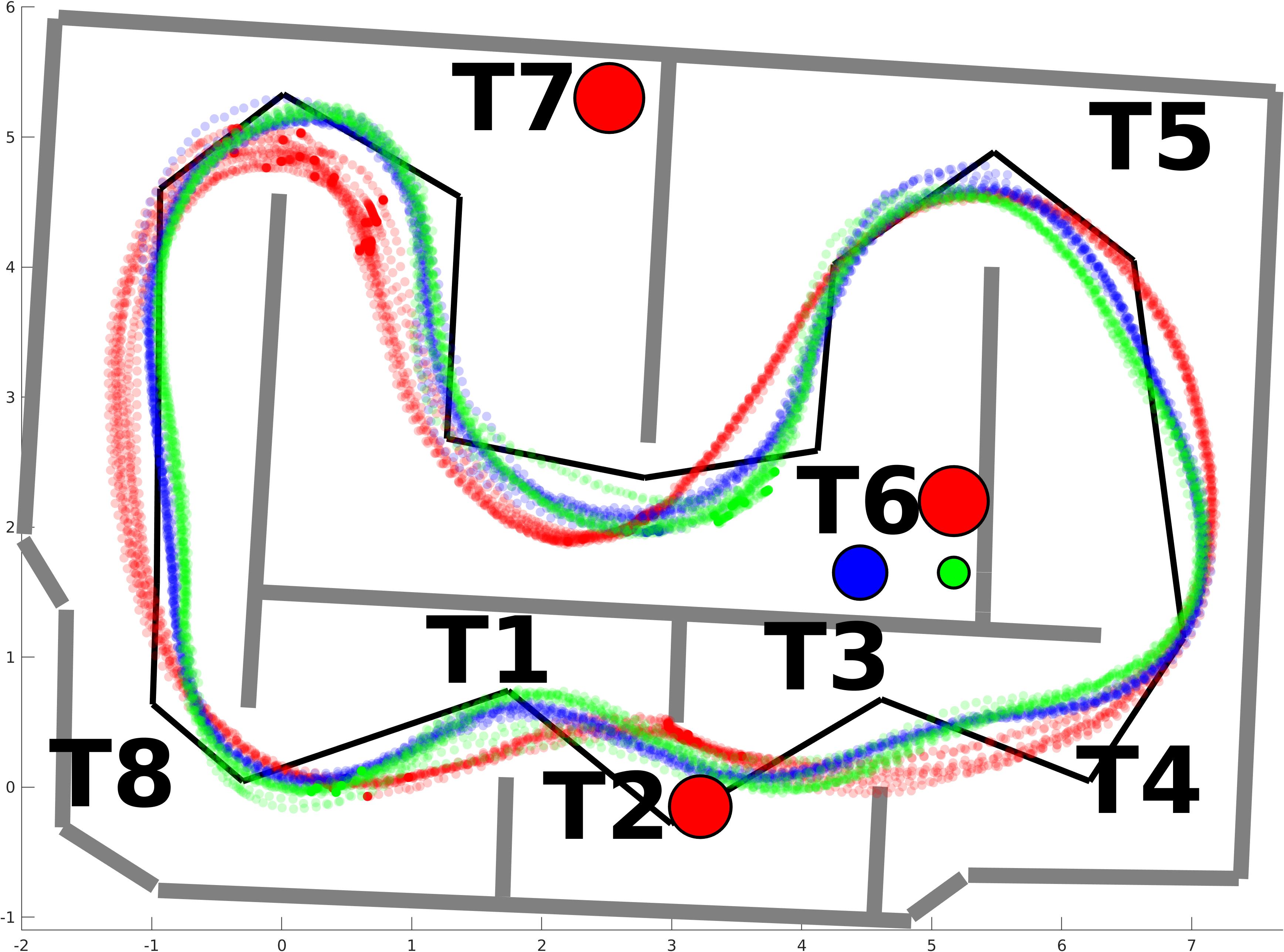}%
\label{fig::1.8}}
\subfloat[1.9m/s]{\includegraphics[width=0.4\columnwidth]{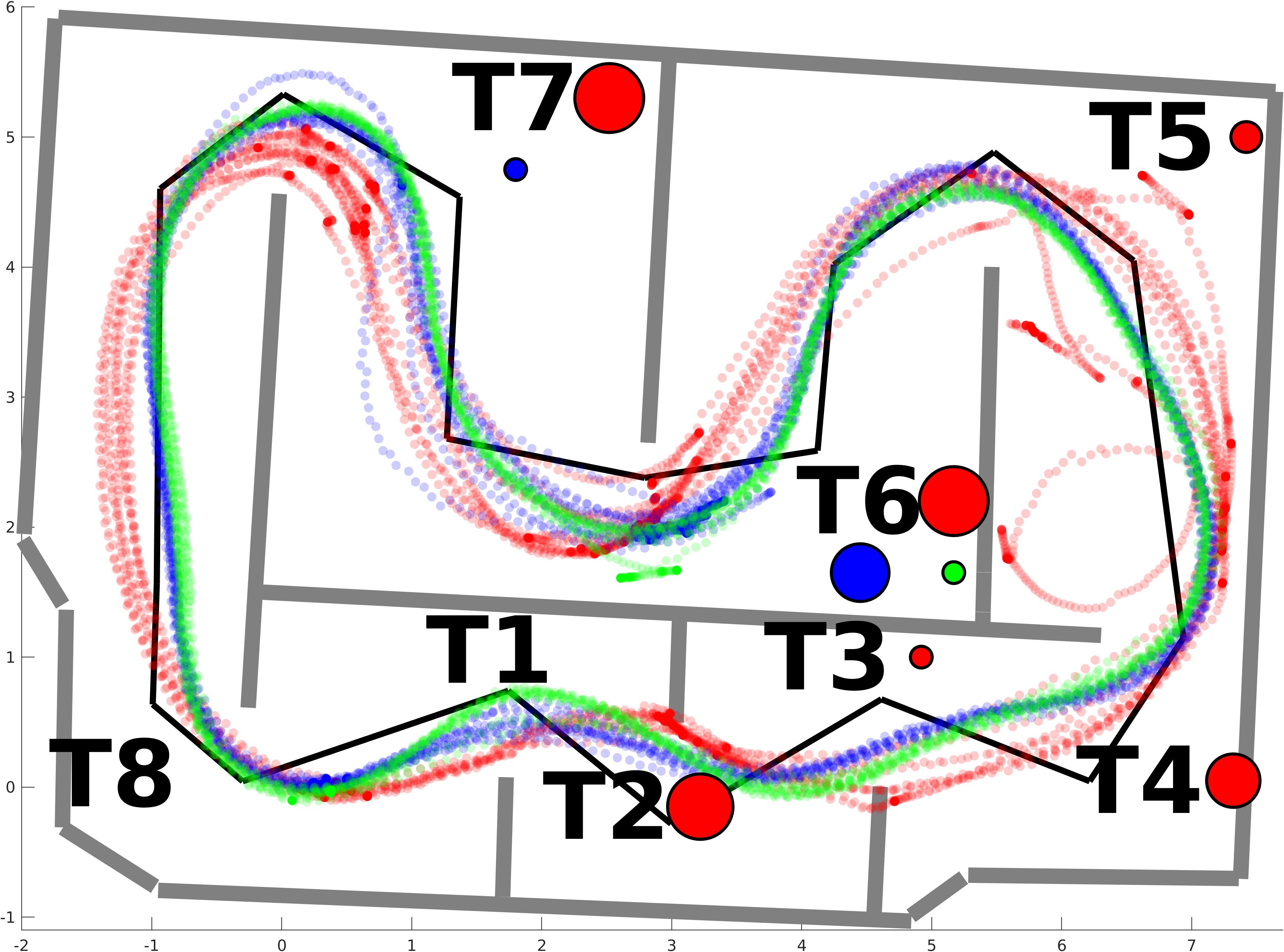}%
\label{fig::1.9}}
\subfloat[2.0m/s]{\includegraphics[width=0.4\columnwidth]{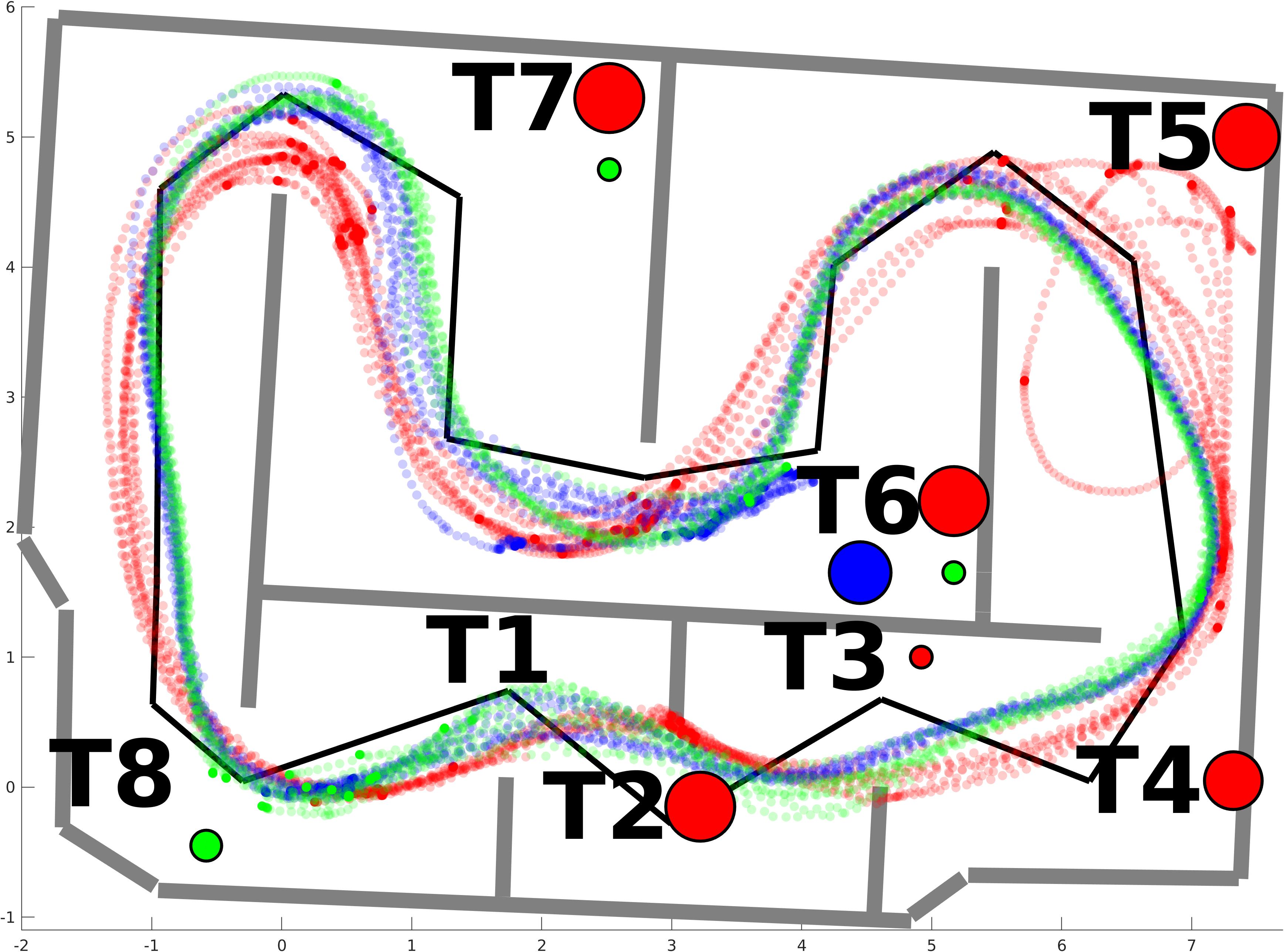}%
\label{fig::2.0}}\\
\subfloat[2.1m/s]{\includegraphics[width=0.4\columnwidth]{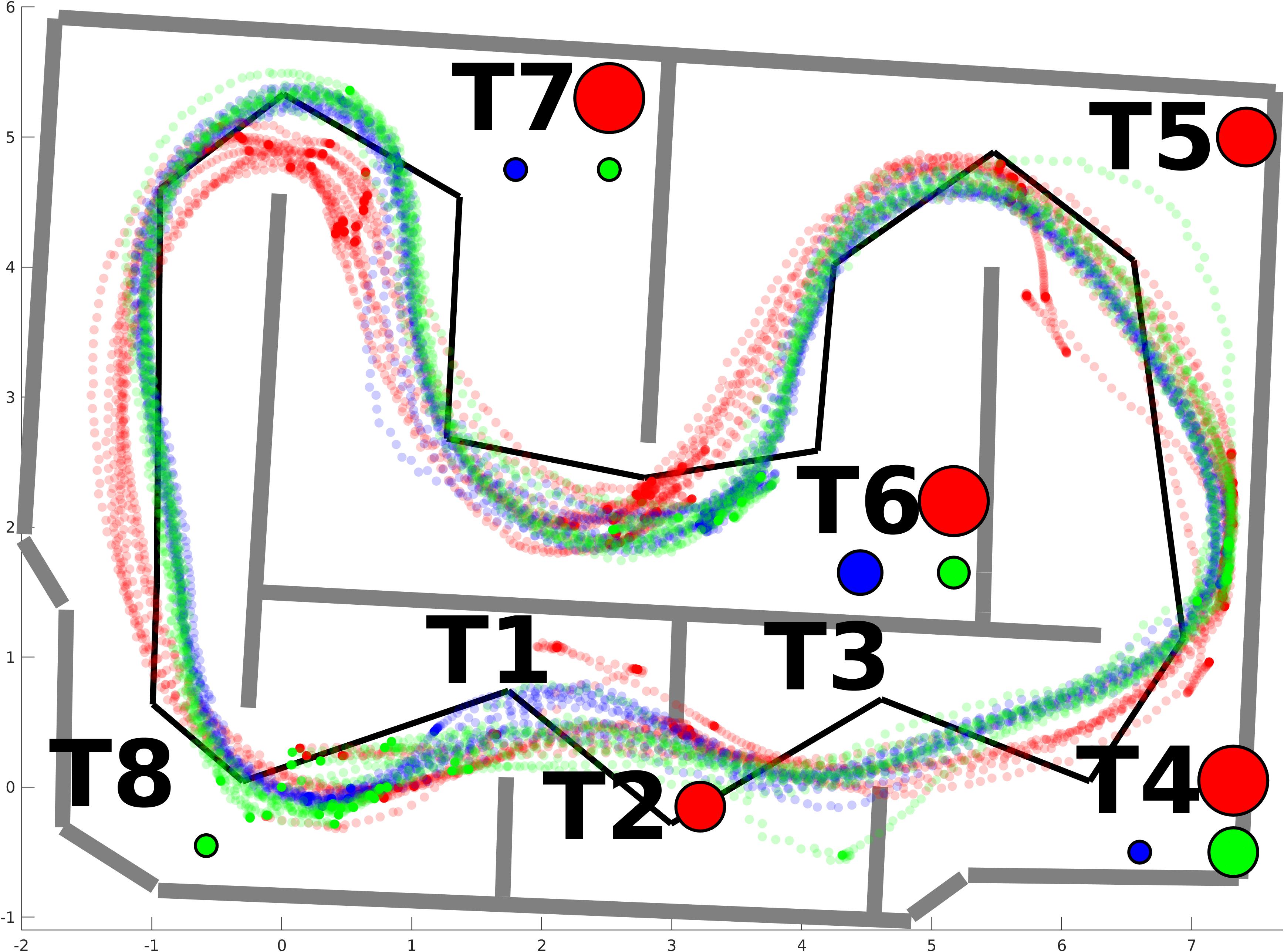}%
\label{fig::2.1}}
\subfloat[2.2m/s]{\includegraphics[width=0.4\columnwidth]{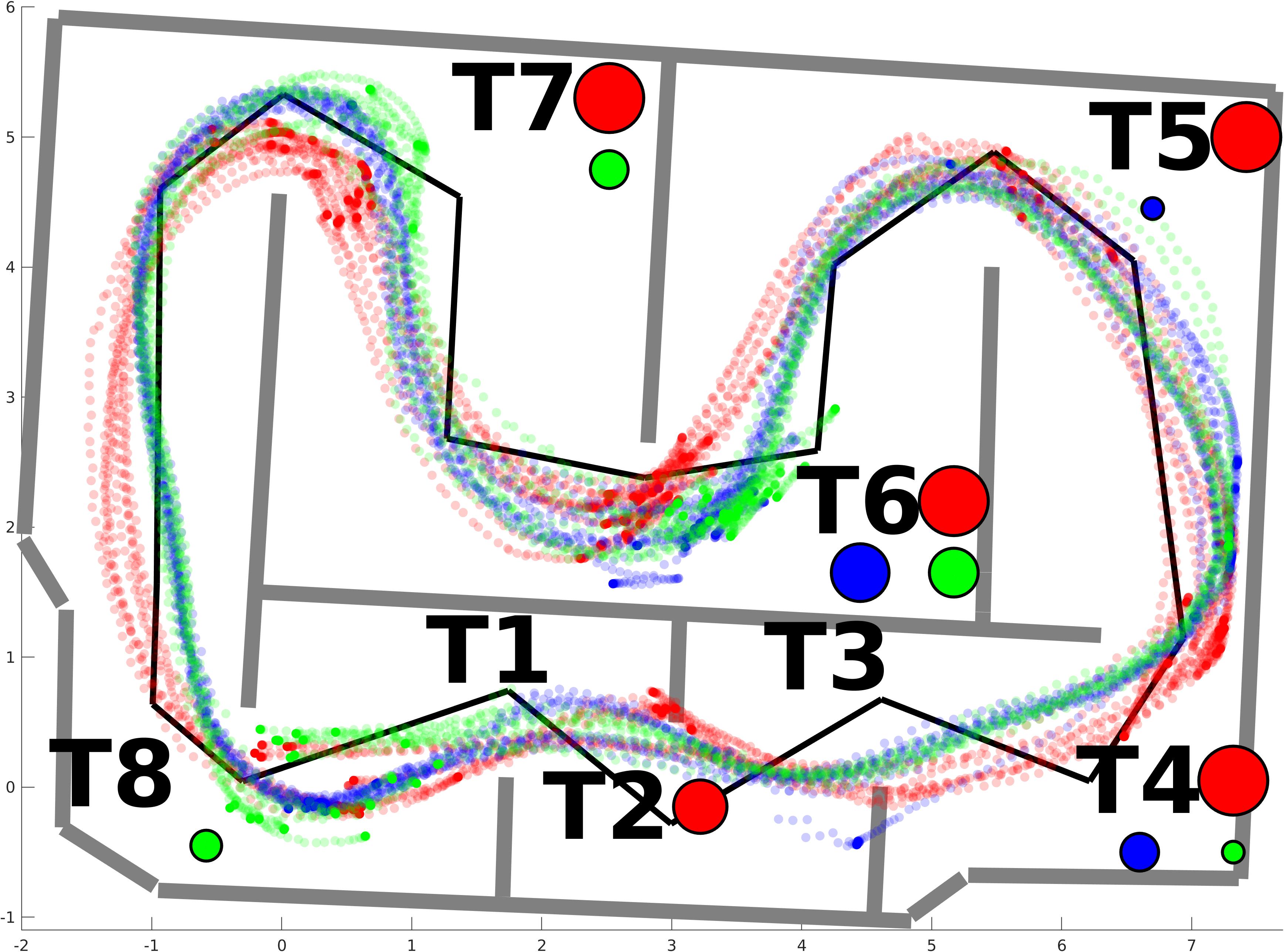}%
\label{fig::2.2}}
\subfloat[2.3m/s]{\includegraphics[width=0.4\columnwidth]{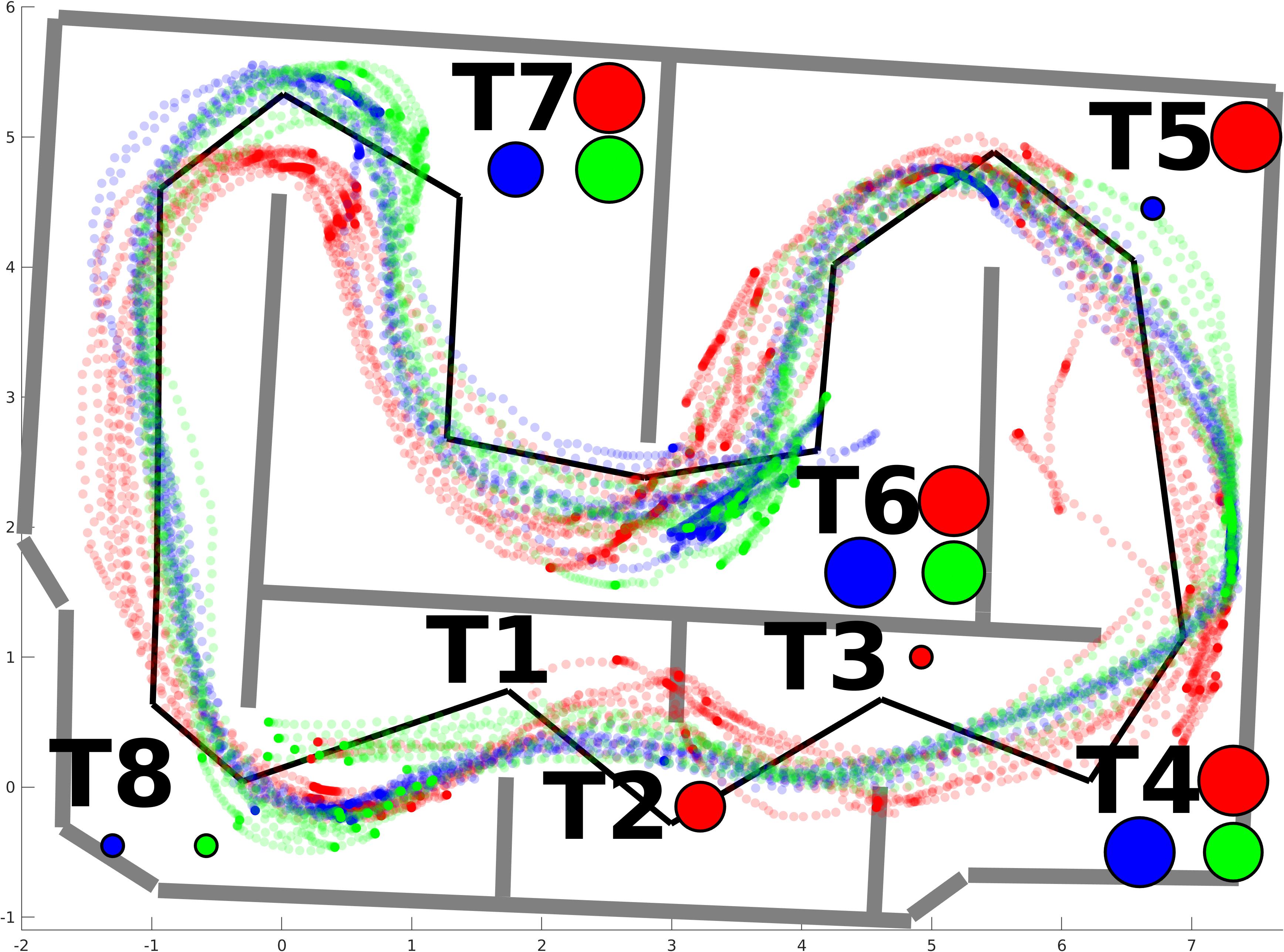}%
\label{fig::2.3}}
\subfloat[2.4m/s]{\includegraphics[width=0.4\columnwidth]{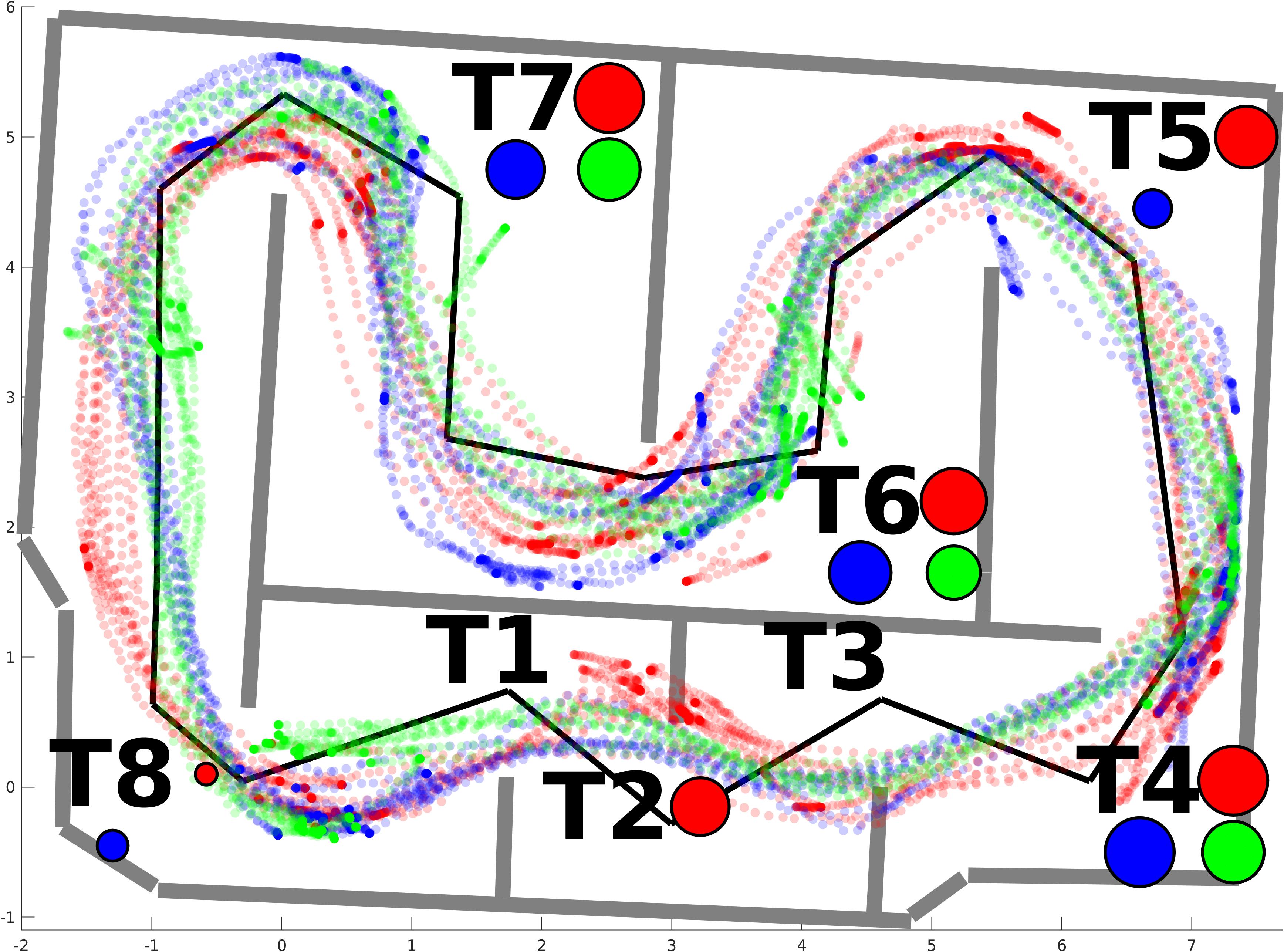}%
\label{fig::2.4}}
\subfloat[2.5m/s]{\includegraphics[width=0.4\columnwidth]{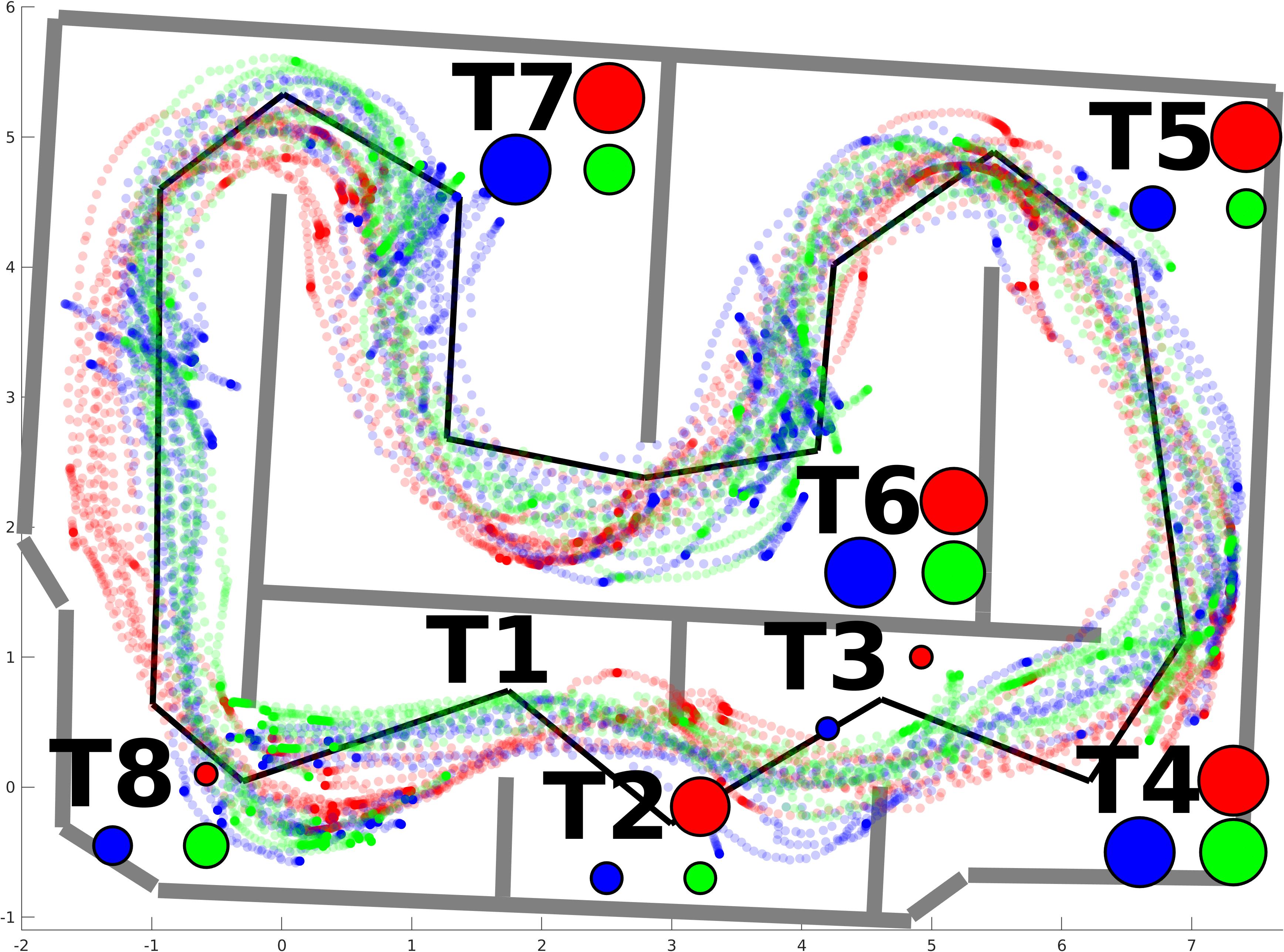}%
\label{fig::2.5}}
\caption{Results of Outdoor Experiments on Seen Terrain: Localized robot positions of all 300 laps are plotted around the pre-defined global path (black line segments) on the map (grey lines). The size of the circles at each turn denotes the number of failures at that turn. Red: baseline $f^+_B(\Delta x, x)$, blue: ablated model $f^+_{A\phi^*}(\Delta x, x)$, green: learned model $f^{+}_{\theta^*}(\Delta x, x, y)$.}
\label{fig::all_data}
\end{figure*}

\begin{figure}
  \centering
  \includegraphics[width=0.9\columnwidth]{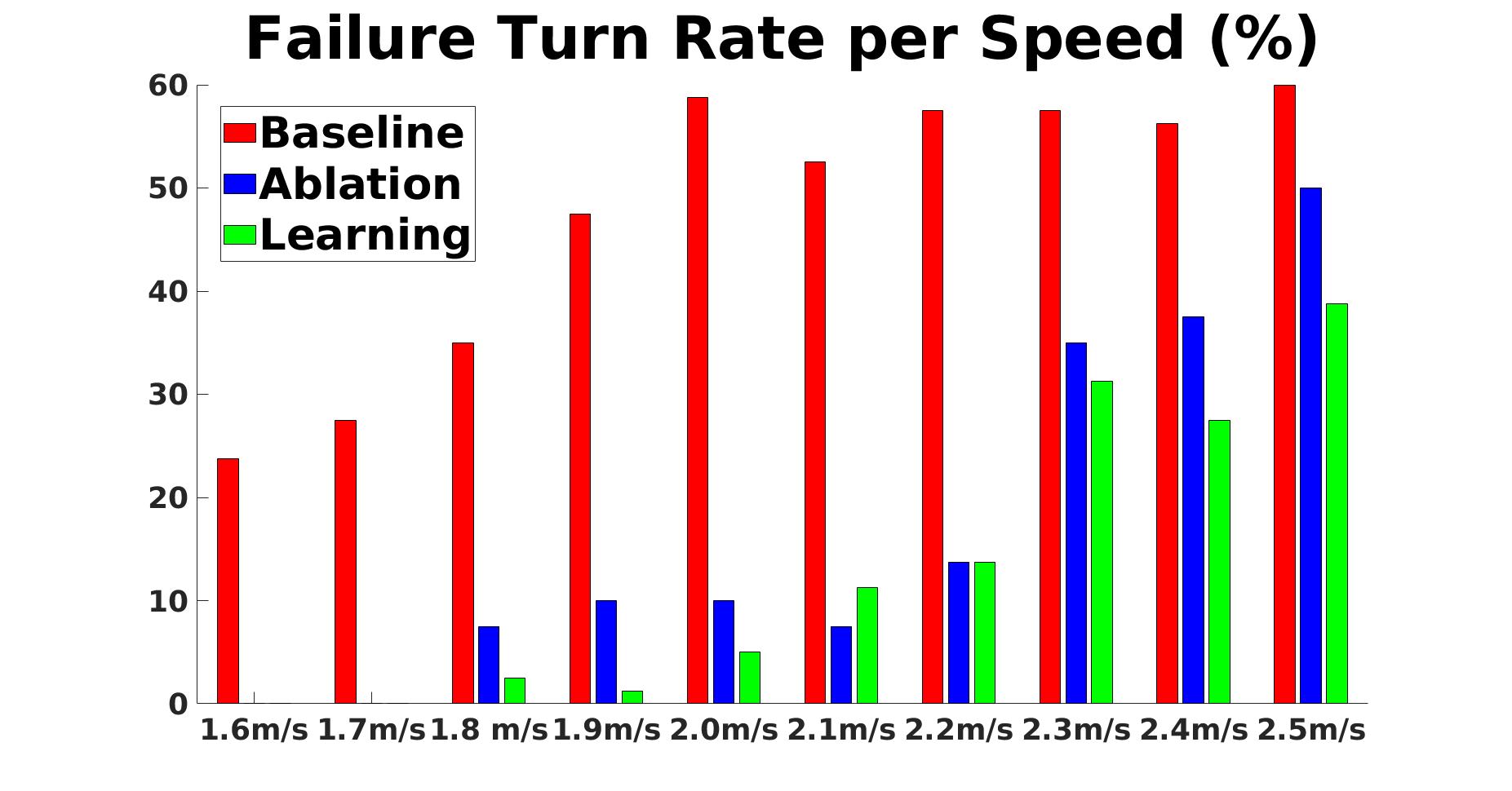}
  \caption{Failure Rate Per Target Speed}
  \label{fig::failure_per_velocity}
\end{figure}

\begin{figure}
  \centering
  \includegraphics[width=0.9\columnwidth]{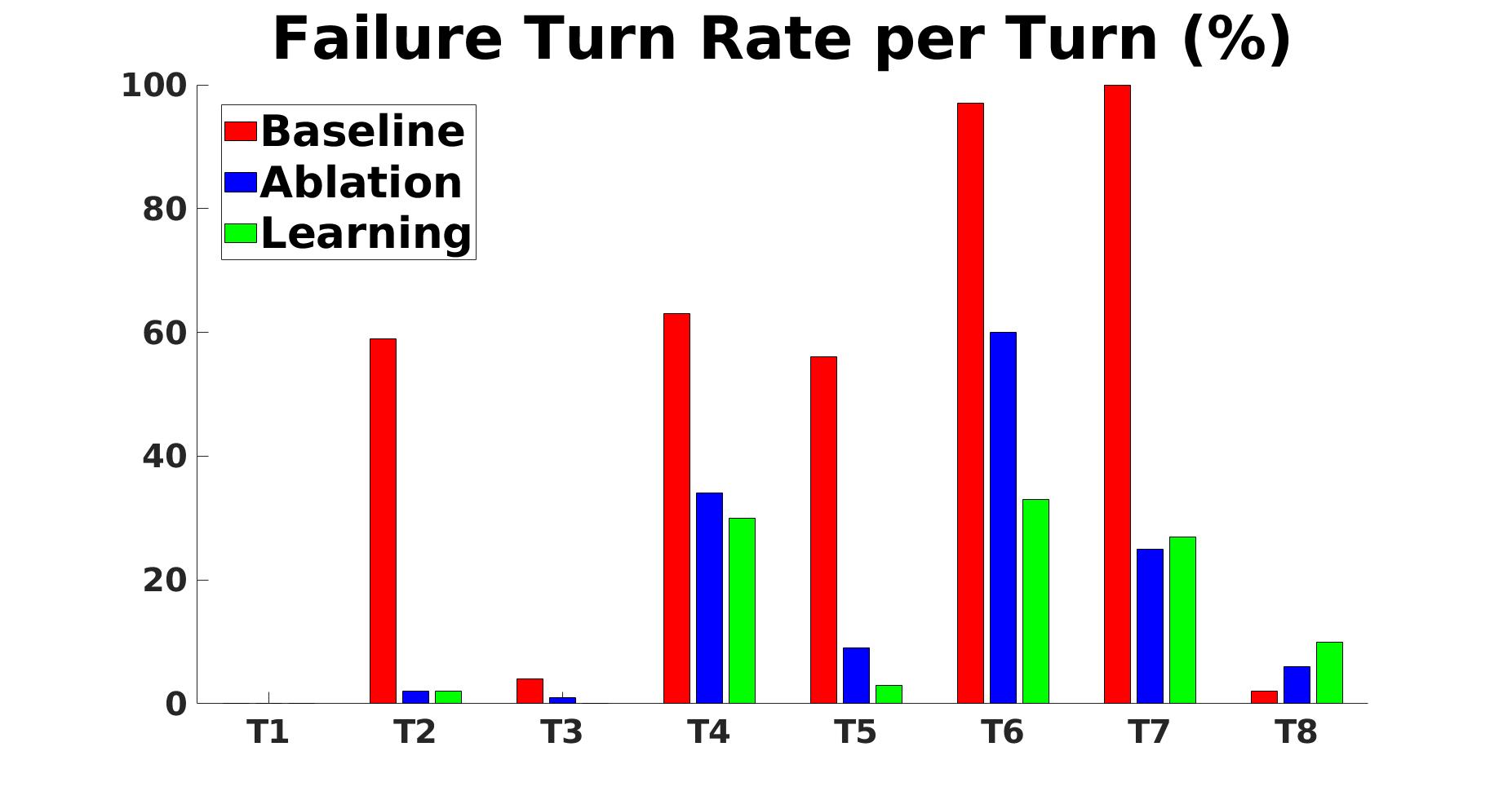}
  \caption{Failure Rate per Turn}
  \label{fig::failure_per_turn}
  \vspace{-10pt}
\end{figure}

\begin{figure*}
\centering
\subfloat[Unseen Test]{\includegraphics[width=0.25581395348\columnwidth]{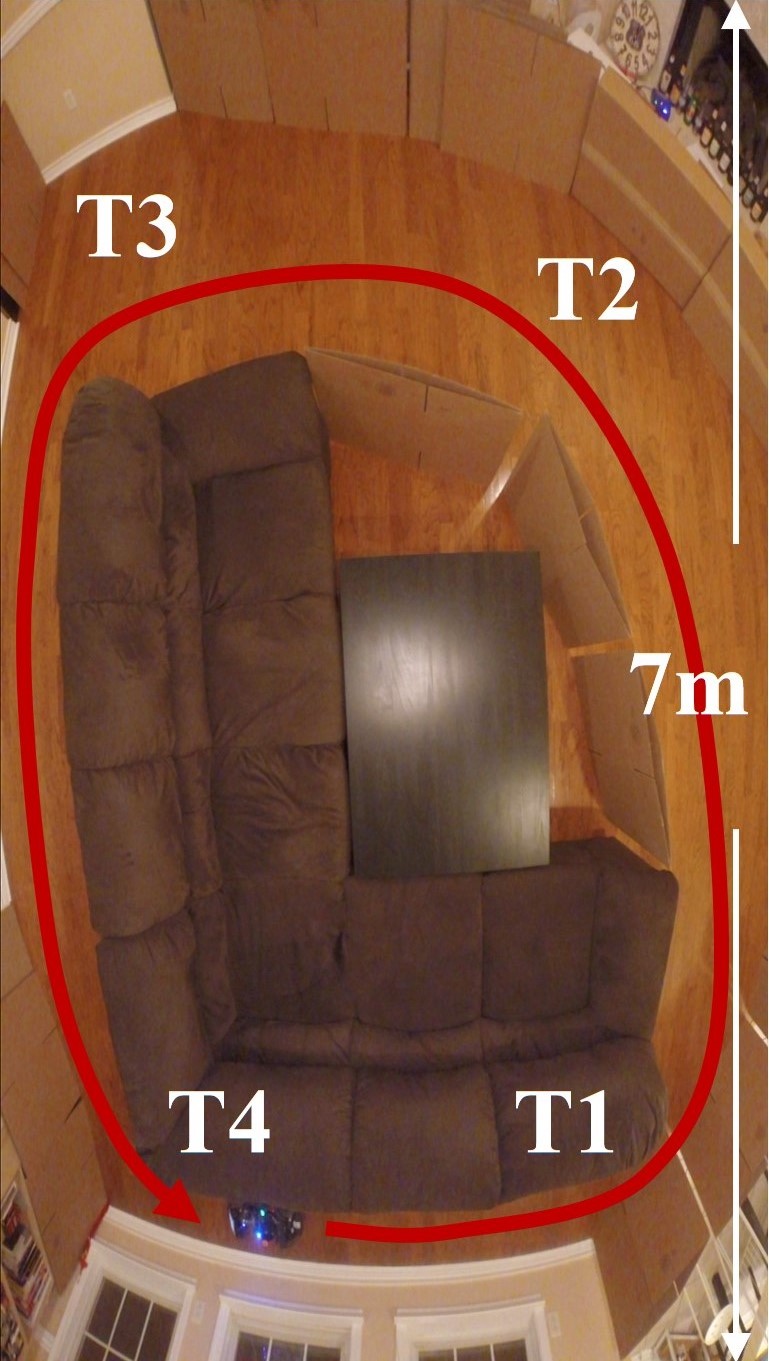}%
\label{fig::living_room}}
\subfloat[2.4m/s]{\includegraphics[width=0.3488372093\columnwidth]{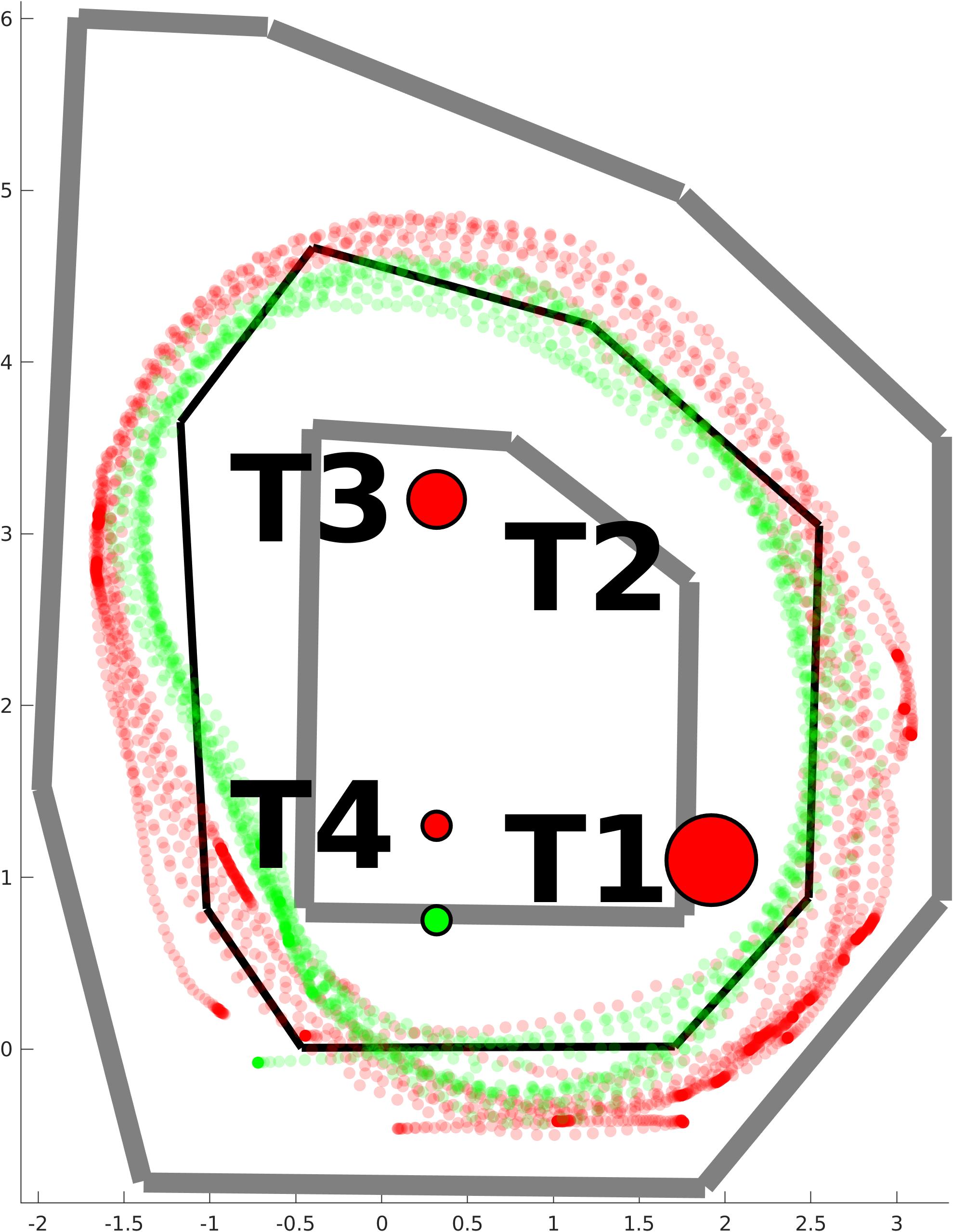}%
\label{fig::2.4_unseen}}
\subfloat[2.5m/s]{\includegraphics[width=0.3488372093\columnwidth]{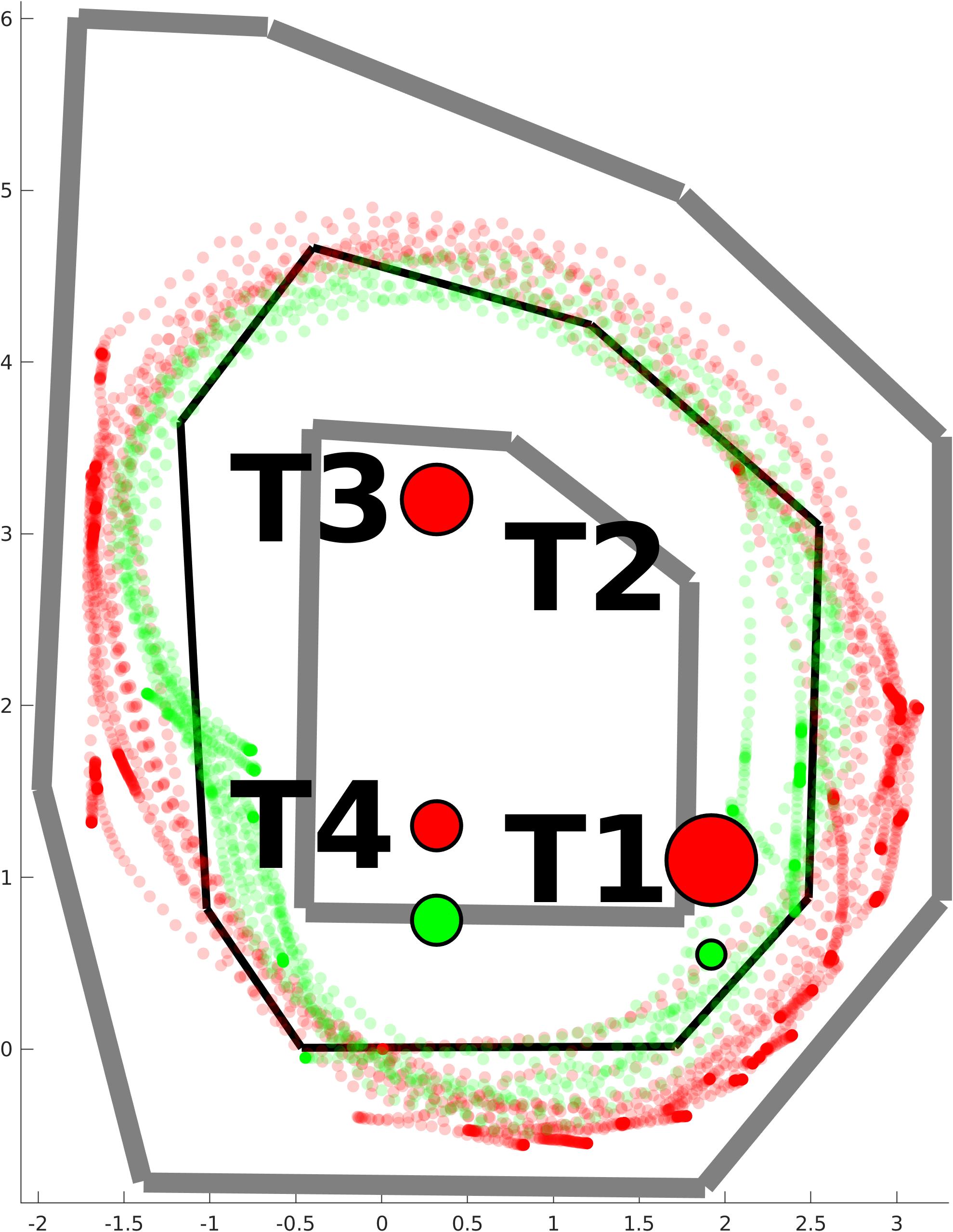}%
\label{fig::2.5_unseen}}
\subfloat[2.6m/s]{\includegraphics[width=0.3488372093\columnwidth]{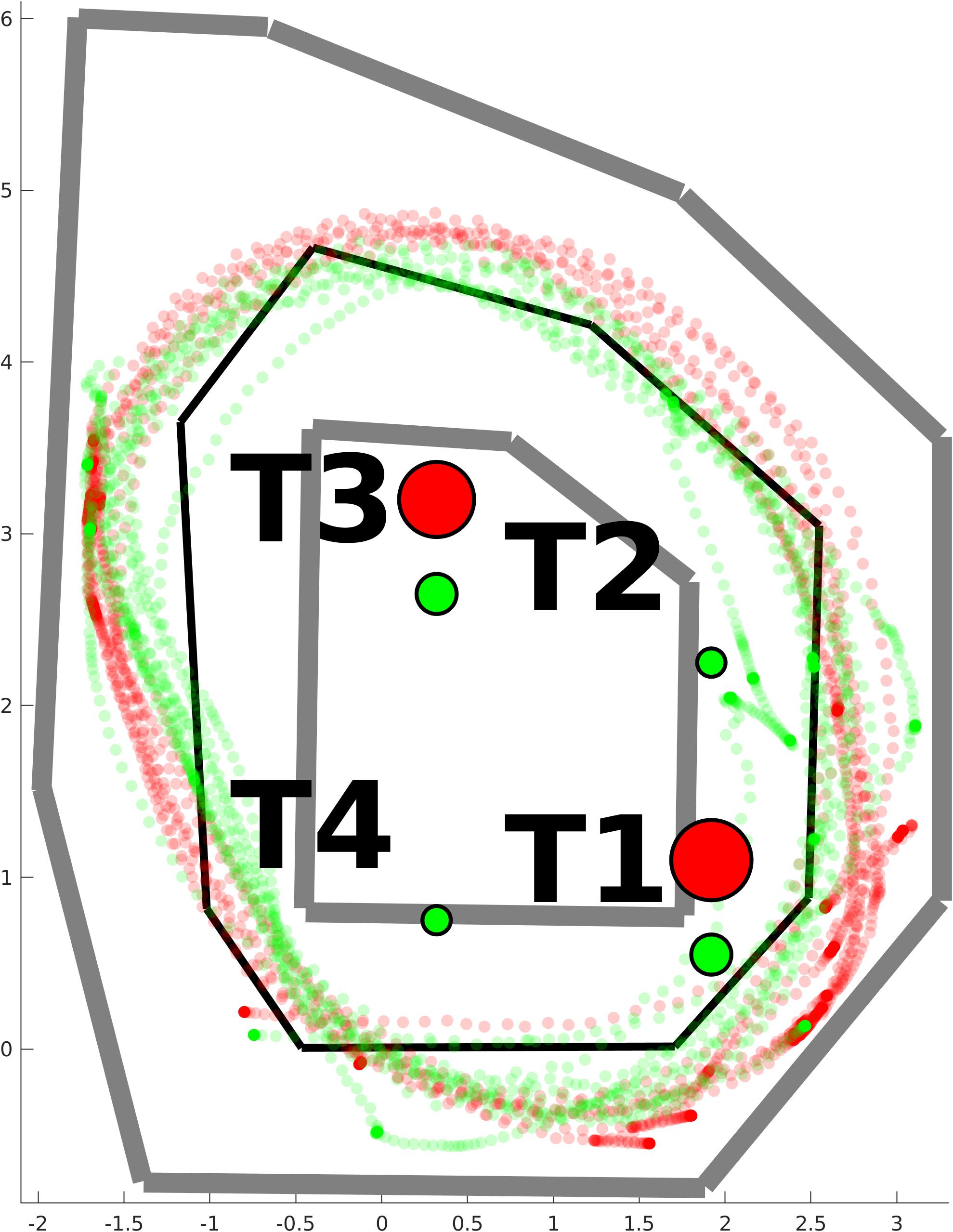}%
\label{fig::2.6_unseen}}
\subfloat[2.7m/s]{\includegraphics[width=0.3488372093\columnwidth]{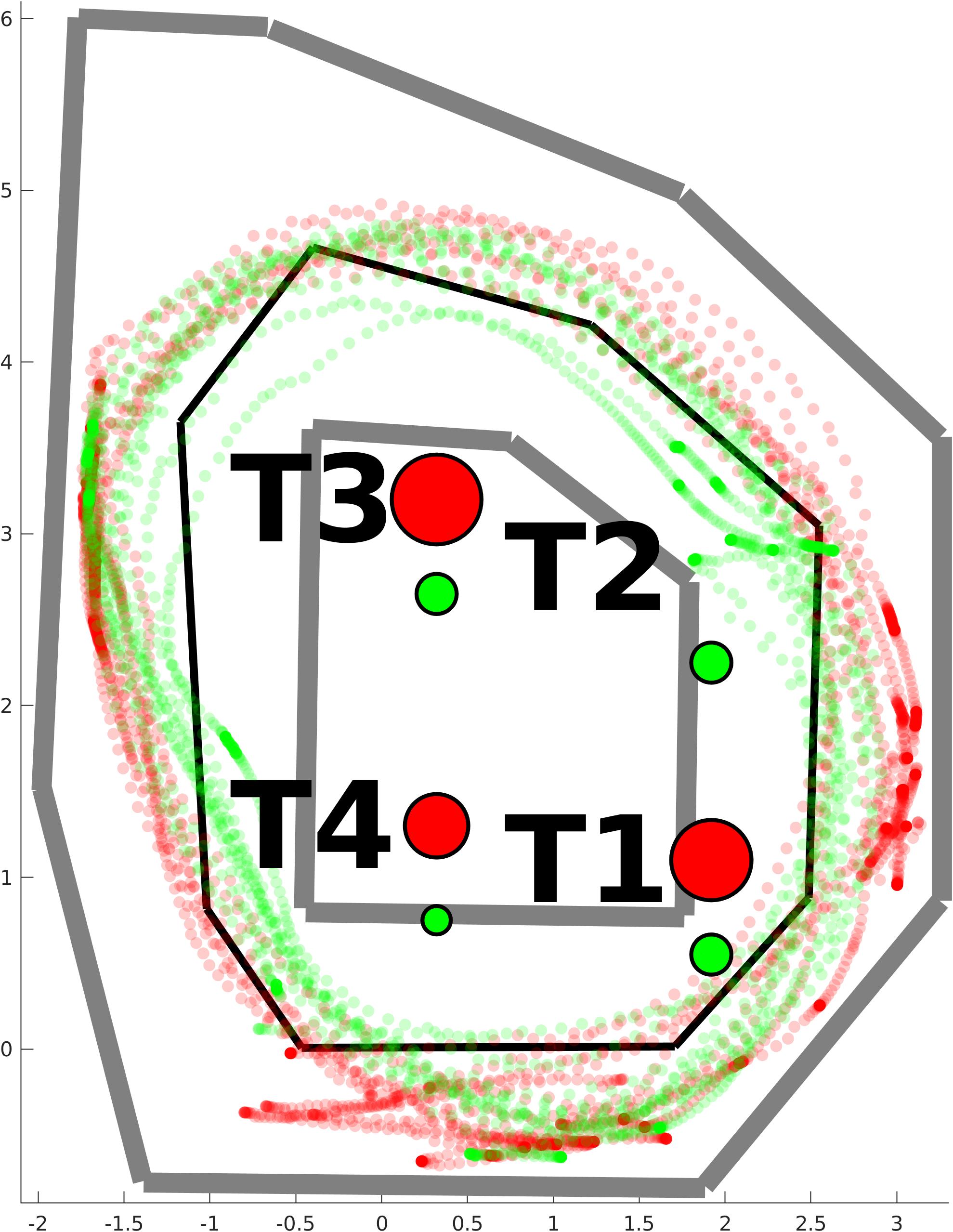}%
\label{fig::2.7_unseen}}
\subfloat[2.8m/s]{\includegraphics[width=0.3488372093\columnwidth]{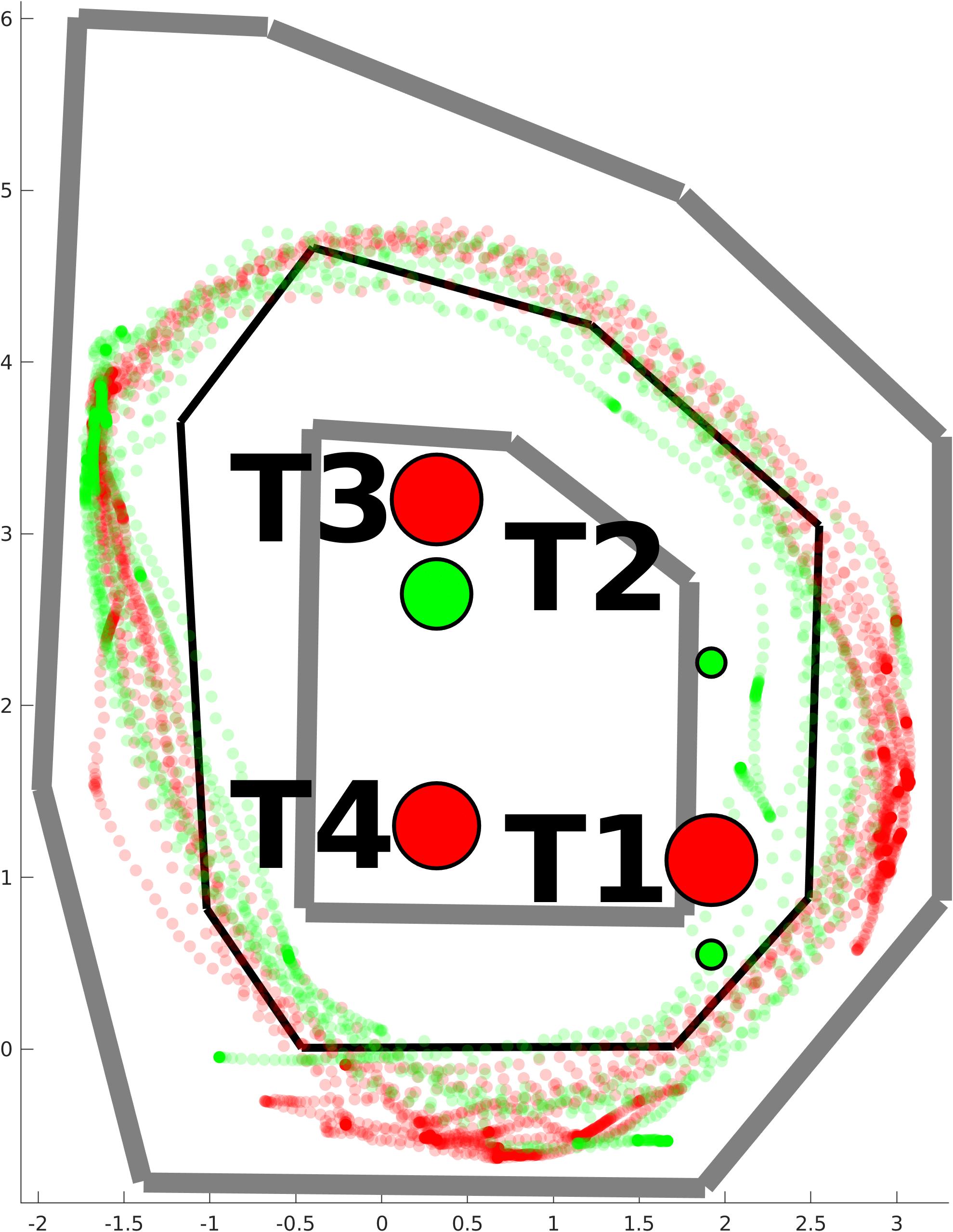}%
\label{fig::2.8_unseen}}
\caption{Experiment Results in Unseen Environment}
\label{fig::all_data_unseen}
\vspace{-10pt}
\end{figure*}

\begin{figure}
  \centering
  \includegraphics[width=0.75\columnwidth]{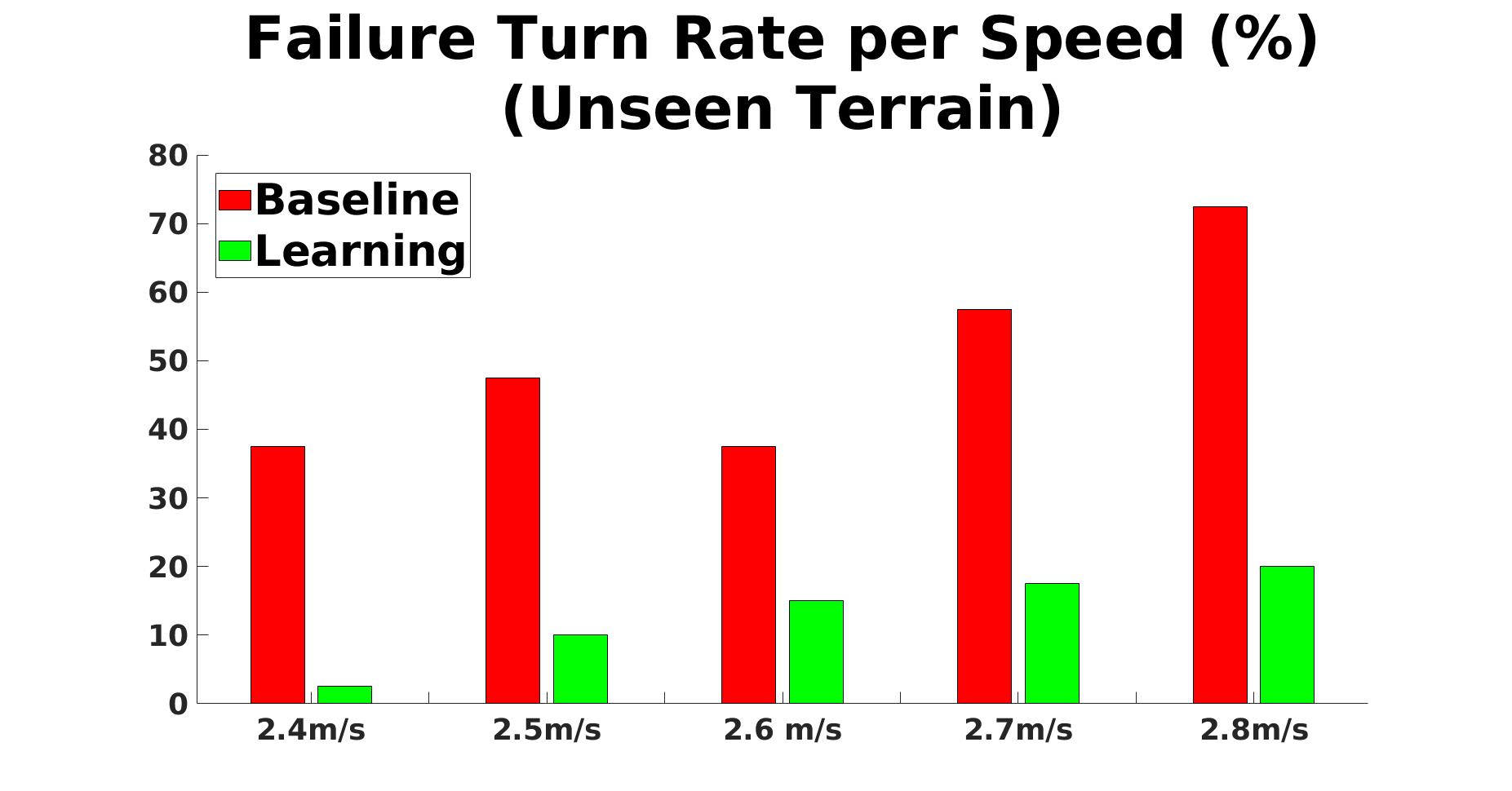}
  \caption{Failure Rate per Target Speed on Unseen Terrain}
  \label{fig::failure_per_velocity_unseen}
  \vspace{-10pt}
\end{figure}

\begin{figure}
  \centering
  \includegraphics[width=0.75\columnwidth]{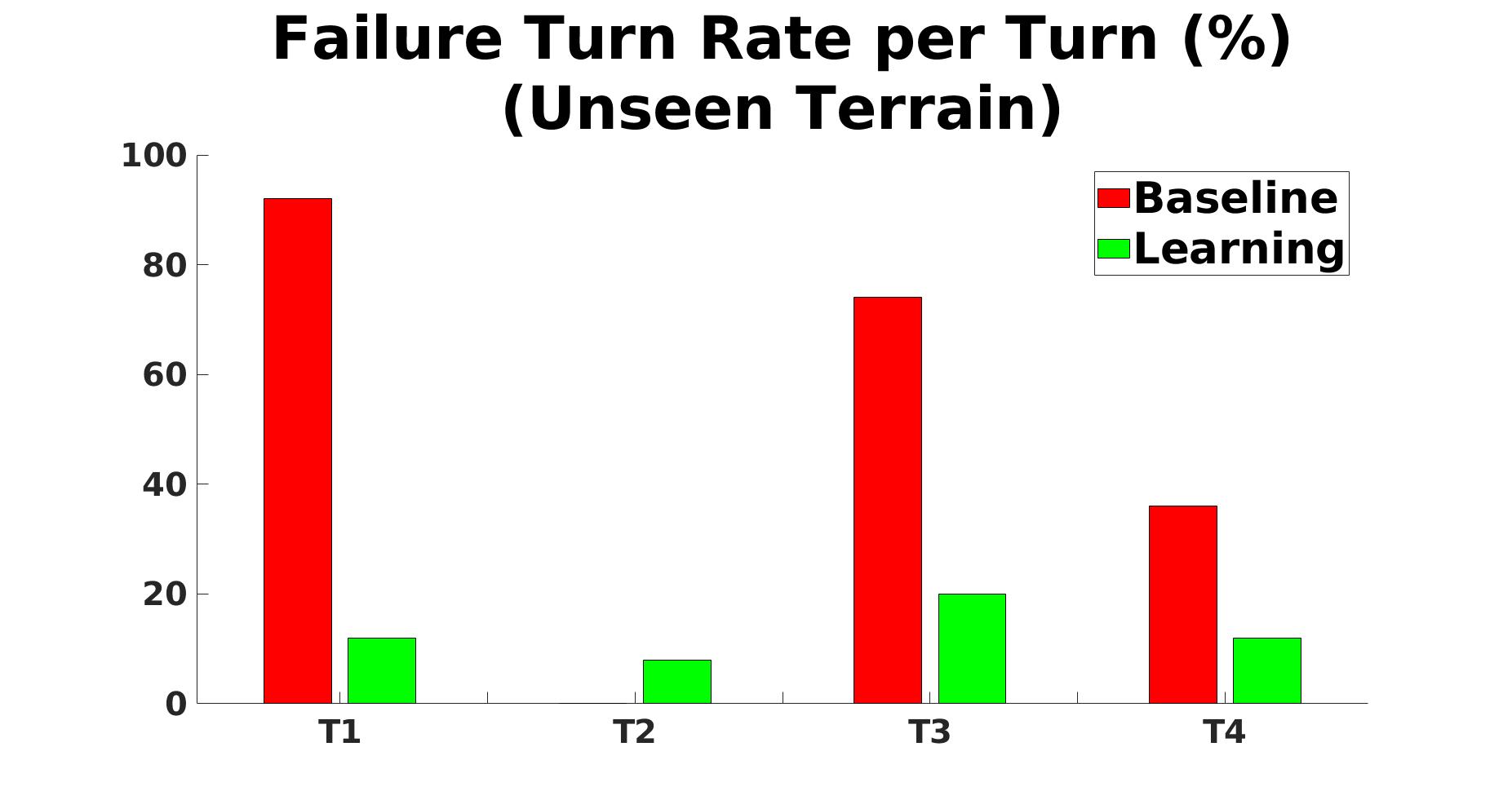}
  \caption{Failure Rate per Turn on Unseen Terrain}
  \label{fig::failure_per_turn_unseen}
  \vspace{-10pt}
\end{figure}

\subsection{Navigation on Seen Terrain}
We first test the inverse kinodynamic model's performance on the same terrain where the training data is collected, but with a different global planner $\Pi_2$. After collecting the training data with $\Pi_1$, an outdoor race track is constructed using plastic panels and wooden posts (Fig. \ref{fig::backyard}). Starting from the origin (robot location in Fig. \ref{fig::backyard}), eight turns are created (T1--T8, $\Pi_2$). While Turn 1, 2, and 3 are relatively gentle \text{left-,} right-, and left-hand turns, Turn 4 and Turn 8 are roughly 90$^{\circ}$ left-hand turns. Turn 5, 6, and 7 are sharp 180$^{\circ}$ left-, right-, and left-hand turns. 

Ten different target speeds (the maximum speed the robot targets at reaching while maintaining safety tolerance to decelerate and avoid potential collisions) are tested, ranging from 1.6m/s to 2.5m/s with 0.1m/s intervals. For the three models, the baseline $f^+_B(\Delta x, x)$, the ablated $f^+_{A\phi^*}(\Delta x, x)$, and learned $f^{+}_{\theta^*}(\Delta x, x, y)$, we repeat ten trials/laps each for statistical significance. A total 300 laps are executed. The localized robot position from ENML are shown in Fig. \ref{fig::all_data} in the subplots corresponding to the target speeds. 

At lower target speeds, the green trajectories by the learned model $f^{+}_{\theta^*}(\Delta x, x, y)$ are much closer to the pre-defined global path, compared to the baseline $f^+_B(\Delta x, x)$, because the latter model fails to consider the world state caused by the unstructured terrain. 
With increasing speed, the robot trajectory becomes more scattered around the global path due to increased stochasticity from vehicle-terrain interaction. But overall speaking, the green trajectories are always closer to the global path than other alternatives. The blue trajectory is generated by the ablated model $f^+_{A\phi^*}(\Delta x, x)$. Like $f^{+}_{\theta^*}$, it learns from actual terrain interactions, but it does not consider the current observation $y$. So $f^+_{A\phi^*}$ is roughly an averaged model over the continuous spectrum of terrain. Therefore, $f^+_{A\phi^*}$ outperforms the baseline $f^+_B$, but underperforms $f^{+}_{\theta^*}$ because it fails to consider the current world state. 

At each turn, the size of the red, blue, and green circles represents the number of failed turns (collision or getting stuck) in the ten attempted turns. Turn 6 and 7 cause a lot of trouble for the baseline even at lower speeds. With increasing speed, more turns cause failure for other models as well, but in general, the baseline fails more frequently at most turns than the ablated and learned models. Fig. \ref{fig::failure_per_velocity} and  Fig. \ref{fig::failure_per_turn} show the percentage of failed turns per target speed and per turn, respectively. In Fig. \ref{fig::failure_per_velocity}, failure rate increases with faster speed, while within each speed, the learned model achieves the lowest failure rate, while the baseline fails most frequently. In Fig. \ref{fig::failure_per_turn}, Turn 6 is the most difficult for all three alternatives, and at most turns, the learned model outperforms the ablation and the baseline. Since Turn 8 is immediately after the terrain change from grass to cement, the robot sometimes oversteers (to compensate for slip on grass) and does not react quickly enough to understeer (for higher friction on cement), causing it to get stuck in a few laps with the learned model. This problem can be addressed by adding forward looking camera to predict future wheel-terrain interaction in future work. The overall success rates of the three models for all turns are shown in the first row in Tab. \ref{tab::success_rates}.

\begin{table}
\centering
\caption{Overall Success Rates of All Speeds and Turns}
\begin{tabular}{cccc}
\toprule
          & Baseline & Ablation & \textbf{Learning}\\ 
\midrule
Seen & 52.4\% & 82.9\% & \textbf{86.9}\% \\
Unseen & 49.5\% & -- & \textbf{87.0}\% \\
\bottomrule
\end{tabular}
\label{tab::success_rates}
\vspace{-12pt}
\end{table}

\subsection{Navigation on Unseen Terrain}

To test that the learned model generalizes to different global planners $\Pi$ and also to unseen terrain, we further conduct an indoor experiment with a different track and global path ($\Pi_3$) on an unseen wooden floor (Fig. \ref{fig::living_room}). Note that $f^{+}_{\theta^*}$ has only seen training data from random exploration on the outdoor terrain (Fig. \ref{fig::backyard}). Since the unseen wooden floor is relatively more consistent and therefore easier to navigate than the outdoor unstructured terrain,\footnote{We speculate that the generalization would not be as good were the model trained indoors (on easy terrain) but applied outdoors.} we increase the navigation target speed to 2.4m/s - 2.8m/s, also with 0.1m/s intervals. The baseline and the learned model are applied with these five different target speeds, ten repetitions each. Fig. \ref{fig::all_data_unseen} shows the results from the 100 laps on the unseen terrain with a different global path. Similar to the results on seen terrain, the learned model produces more concentrated and also closer robot trajectories to the global path to be tracked. As shown in Fig. \ref{fig::failure_per_velocity_unseen} and \ref{fig::failure_per_turn_unseen}, the learned model also outperforms the baseline in terms of failure rate at all target speeds and in most turns (except Turn 2). The overall success rates of the baseline and learned model are shown in the second row in Tab. \ref{tab::success_rates}.

\section{Conclusions}
\label{sec::conclusions}

In this paper, we present a data-driven approach to learn an inverse kinodynamic model for accurate high-speed navigation on unstructured terrain. To capture the elusive and stochastic world state caused by vehicle-terrain interaction at different high speeds, we use an inertia-based observation embedding as an input to the learned inverse kinodynamic function. This approach is tested on a physical robot on seen and unseen terrain with different global plans at different high speeds. The experimental results show that the learned model can significantly outperform an ideal baseline model without consideration of world state. Our ablation study also shows our observation embedding is useful to enable fast and accurate off-road navigation on unstructured terrain. For future work, better ground truth linear velocity estimation needs to be investigated: in addition to wheel odometry alone, other sources of perception, e.g., vision, point cloud, and/or inertia, can be leveraged. Better linear velocity estimation can account for significant wheel slippage on more challenging terrain, e.g., on ice, and enable even faster navigation. Adding vision-based observation also has the potential to enable the robot to prepare for future interactions, e.g., to reduce the failures at Turn 8. Another interesting direction to investigate in the future is generalization from easier to harder environments. 

\section*{ACKNOWLEDGMENTS}
This work has taken place in the Learning Agents Research Group (LARG) and Autonomous Mobile Robotics Laboratory (AMRL) at UT Austin.  LARG research is supported in part by NSF (CPS-1739964, IIS-1724157, NRI-1925082), ONR (N00014-18-2243), FLI (RFP2-000), ARO (W911NF-19-2-0333), DARPA, Lockheed Martin, GM, and Bosch. 
AMRL research is supported in part by NSF (CAREER-2046955, IIS-1954778, SHF-2006404), ARO (W911NF-19-2-0333), DARPA (HR001120C0031), Amazon, JP Morgan, and Northrop Grumman Mission Systems.
Peter Stone serves as the Executive Director of Sony AI America and receives financial compensation for this work.  The terms of this arrangement have been reviewed and approved by the University of Texas at Austin in accordance with its policy on objectivity in research.

\bibliographystyle{IEEEtran}
\bibliography{IEEEabrv,references}

\end{document}